\documentclass{article}
\PassOptionsToPackage{numbers, compress}{natbib}

\usepackage[final]{neurips_2021}

\usepackage[utf8]{inputenc} 
\usepackage[T1]{fontenc}    
\usepackage[colorlinks=true,linkcolor=blue,urlcolor=blue,citecolor=purple]{hyperref}       
\usepackage{url}            
\usepackage{booktabs}       
\usepackage{amsfonts}       
\usepackage{nicefrac}       
\usepackage{microtype}      
\usepackage{xcolor}         
\usepackage{mathtools}
\usepackage{etoolbox}
\usepackage{amssymb}
\usepackage{bm}
\usepackage{float}
\usepackage{graphicx}
\usepackage{booktabs,multirow,array}
\usepackage{fullpage}
\usepackage{siunitx}

\usepackage{listings}
\usepackage[utf8]{inputenc}
\usepackage[english]{babel}
\definecolor{maroon}{cmyk}{0, 0.87, 0.68, 0.32}
\definecolor{halfgray}{gray}{0.55}
\definecolor{ipython_frame}{RGB}{207, 207, 207}
\definecolor{ipython_bg}{RGB}{247, 247, 247}
\definecolor{ipython_red}{RGB}{186, 33, 33}
\definecolor{ipython_green}{RGB}{0, 128, 0}
\definecolor{ipython_cyan}{RGB}{64, 128, 128}
\definecolor{ipython_purple}{RGB}{170, 34, 255}

\colorlet{mygray}{black!30}
\colorlet{mygreen}{green!60!blue}
\colorlet{mymauve}{red!60!blue}
\lstdefinelanguage{commandline}{
	basicstyle=\ttfamily,
	columns=fullflexible,
	breakatwhitespace=false,
	breaklines=true,
	captionpos=b,
	commentstyle=\color{mygreen},
	extendedchars=true,
	frame=single,
	keepspaces=true,
	keywordstyle=\color{blue},
	language=c++,
	numbers=none,
	numbersep=5pt,
	numberstyle=\tiny\color{blue},
	rulecolor=\color{mygray},
	showspaces=false,
	showtabs=false,
	stepnumber=5,
	stringstyle=\color{mymauve},
	tabsize=3,
	title=\lstname
}
\lstdefinelanguage{iPython}{
morekeywords={access,and,break,class,continue,def,del,elif,else,except,exec,finally,for,from,global,if,import,in,is,lambda,not,or,pass,print,raise,return,try,while},%
morekeywords=[2]{abs,all,any,basestring,bin,bool,bytearray,callable,chr,classmethod,cmp,compile,complex,delattr,dict,dir,divmod,enumerate,eval,execfile,file,filter,float,format,frozenset,getattr,globals,hasattr,hash,help,hex,id,input,int,isinstance,issubclass,iter,len,list,locals,long,map,max,memoryview,min,next,object,oct,open,ord,pow,property,range,raw_input,reduce,reload,repr,reversed,round,set,setattr,slice,sorted,staticmethod,str,sum,super,tuple,type,unichr,unicode,vars,xrange,zip,apply,buffer,coerce,intern},%
sensitive=true,%
morecomment=[l]\#,%
morestring=[b]',%
morestring=[b]",%
morestring=[s]{'''}{'''},
morestring=[s]{"""}{"""},
morestring=[s]{r'}{'},
morestring=[s]{r"}{"},%
morestring=[s]{r'''}{'''},%
morestring=[s]{r"""}{"""},%
morestring=[s]{u'}{'},
morestring=[s]{u"}{"},%
morestring=[s]{u'''}{'''},%
morestring=[s]{u"""}{"""},%
literate=
	{á}{{\'a}}1 {é}{{\'e}}1 {í}{{\'i}}1 {ó}{{\'o}}1 {ú}{{\'u}}1
{Á}{{\'A}}1 {É}{{\'E}}1 {Í}{{\'I}}1 {Ó}{{\'O}}1 {Ú}{{\'U}}1
{à}{{\`a}}1 {è}{{\`e}}1 {ì}{{\`i}}1 {ò}{{\`o}}1 {ù}{{\`u}}1
{À}{{\`A}}1 {È}{{\'E}}1 {Ì}{{\`I}}1 {Ò}{{\`O}}1 {Ù}{{\`U}}1
{ä}{{\"a}}1 {ë}{{\"e}}1 {ï}{{\"i}}1 {ö}{{\"o}}1 {ü}{{\"u}}1
{Ä}{{\"A}}1 {Ë}{{\"E}}1 {Ï}{{\"I}}1 {Ö}{{\"O}}1 {Ü}{{\"U}}1
{â}{{\^a}}1 {ê}{{\^e}}1 {î}{{\^i}}1 {ô}{{\^o}}1 {û}{{\^u}}1
{Â}{{\^A}}1 {Ê}{{\^E}}1 {Î}{{\^I}}1 {Ô}{{\^O}}1 {Û}{{\^U}}1
{œ}{{\oe}}1 {Œ}{{\OE}}1 {æ}{{\ae}}1 {Æ}{{\AE}}1 {ß}{{\ss}}1
{ç}{{\c c}}1 {Ç}{{\c C}}1 {ø}{{\o}}1 {å}{{\r a}}1 {Å}{{\r A}}1
{€}{{\EUR}}1 {£}{{\pounds}}1
{^}{{{\color{ipython_purple}\^{}}}}1
{=}{{{\color{ipython_purple}=}}}1
{+}{{{\color{ipython_purple}+}}}1
{*}{{{\color{ipython_purple}$^\ast$}}}1
{/}{{{\color{ipython_purple}/}}}1
{+=}{{{+=}}}1
{-=}{{{-=}}}1
{*=}{{{$^\ast$=}}}1
{/=}{{{/=}}}1,
literate=
	*{-}{{{\color{ipython_purple}-}}}1
{?}{{{\color{ipython_purple}?}}}1,
identifierstyle=\color{black}\ttfamily,
commentstyle=\color{ipython_cyan}\ttfamily,
stringstyle=\color{ipython_red}\ttfamily,
keepspaces=true,
showspaces=false,
showstringspaces=false,
rulecolor=\color{ipython_frame},
frame=single,
frameround={t}{t}{t}{t},
framexleftmargin=6mm,
numbers=left,
numberstyle=\tiny\color{halfgray},
backgroundcolor=\color{ipython_bg},
basicstyle=\scriptsize,
keywordstyle=\color{ipython_green}\ttfamily,
}
\lstdefinestyle{DOS}
{
	backgroundcolor=\color{black},
	basicstyle=\scriptsize\color{white}\ttfamily
}
\usepackage{tikz-cd}
\usetikzlibrary{calc}
\usetikzlibrary{decorations.pathmorphing}

\tikzset{curve/.style={settings={#1},to path={(\tikztostart)
					.. controls ($(\tikztostart)!\pv{pos}!(\tikztotarget)!\pv{height}!270:(\tikztotarget)$)
					and ($(\tikztostart)!1-\pv{pos}!(\tikztotarget)!\pv{height}!270:(\tikztotarget)$)
					.. (\tikztotarget)\tikztonodes}},
	settings/.code={\tikzset{quiver/.cd,#1}
			\def\pv##1{\pgfkeysvalueof{/tikz/quiver/##1}}},
	quiver/.cd,pos/.initial=0.35,height/.initial=0}

\tikzset{tail reversed/.code={\pgfsetarrowsstart{tikzcd to}}}
\tikzset{2tail/.code={\pgfsetarrowsstart{Implies[reversed]}}}
\tikzset{2tail reversed/.code={\pgfsetarrowsstart{Implies}}}
\tikzset{no body/.style={/tikz/dash pattern=on 0 off 1mm}}

\newcommand\quotient[2]{
	\mathchoice
	{
		\text{\raise1ex\hbox{$#1$}\Big/\lower1ex\hbox{$#2$}}%
	}
	{
		#1\,/\,#2
	}
	{
		#1\,/\,#2
	}
	{
		#1\,/\,#2
	}
}

\usepackage{physics}
\usepackage{complexity}

\usepackage{stmaryrd} 

\usepackage{bm}
\usepackage{bbm}

\usepackage{import}
\usepackage{xifthen}
\usepackage{pdfpages}
\usepackage{transparent}

\usepackage{appendix}

\makeatletter
\patchcmd{\hyper@makecurrent}{%
	\ifx\Hy@param\Hy@chapterstring
		\let\Hy@param\Hy@chapapp
	\fi}{%
	\iftoggle{inappendix}{
		\@checkappendixparam{chapter}%
		\@checkappendixparam{section}%
		\@checkappendixparam{subsection}%
		\@checkappendixparam{subsubsection}%
		\@checkappendixparam{paragraph}%
		\@checkappendixparam{subparagraph}%
	}{}%
}{}{\errmessage{failed to patch}}

\newcommand*{\@checkappendixparam}[1]{%
	\def\@checkappendixparamtmp{#1}%
	\ifx\Hy@param\@checkappendixparamtmp
		\let\Hy@param\Hy@appendixstring
	\fi
}
\makeatletter

\newtoggle{inappendix}
\togglefalse{inappendix}

\apptocmd{\appendix}{\toggletrue{inappendix}}{}{\errmessage{failed to patch}}
\apptocmd{\subappendices}{\toggletrue{inappendix}}{}{\errmessage{failed to patch}}

\title{Travel the Same Path: A Novel TSP Solving Strategy}

\author{%
  Pingbang Hu\\
  Department of Computer Science\\
  University of Michigan\\
  \texttt{pbb@umich.edu} \\
}

\begin{document}

\maketitle

\begin{abstract}
	In this paper, we provide a novel strategy for solving Traveling Salesman Problem, which is a famous combinatorial optimization problem studied intensely in the TCS community.
	In particular, we consider the imitation learning framework, which helps a deterministic algorithm making good choices whenever it needs to, resulting in a speed up while maintaining the exactness of the solution without suffering from the unpredictability and a potential large deviation. 
 
 Furthermore, we demonstrate a strong generalization ability of a graph neural network trained under the imitation learning framework. Specifically, the model is capable of solving a large instance of TSP faster than the baseline while has only seen small TSP instances when training.

\end{abstract}

\section{Introduction}
The traveling salesman problem (TSP) can be described as follows: given a list of cities and the distances between each pair of cities, find the
shortest route possible that visits each city \emph{exactly once} then returns to the origin city.
Specifically, given an undirected weighted graph \(\mathcal{G} = (\mathcal{E}, \mathcal{V})\), with an ordered pair of nodes set \(\mathcal{E}\)
and an edge set \(\mathcal{V}\subseteq \mathcal{E}\times\mathcal{E}\) where \(\mathcal{G}\) is equipped with \emph{spatial structure}. This means that
each edge between nodes will have different weights and each node will have its coordinates, we want to find a simple cycle that visits every node exactly
once while having the smallest cost.

We will utilize GCNN (Graph Convolutional Neural Network), a particular kind of GNN, together with imitation learning to solve TSP in an interesting
and inspiring way. In particular, we focus on the generalization ability of models trained on small-sized problem instances.\footnote{The code is
	available at \url{https://github.com/sleepymalc/Travel-the-Same-Path}.}

\section{Related Works}
There has already been extensive work done to optimize TSP solvers both theoretically and practically. We have done extensive research into other solvers;
the papers most relevant to our project are summarized below.

\paragraph{Transformer Network for TSP~\cite{Bresson2021TheTN}.}
The main focus of this paper is to detail the application of deep reinforcement learning reapplied to a Transformer architecture originally created for
Natural Language Processing (NLP). Unlike our proposed model, this solver does not solve TSP exactly but instead learns heuristics that have very low
error rates (0.004\% for TSP50 and 0.39\% for TSP100). These heuristics can run over a TSP problem much faster than a traditional solver while still
achieving similar results.

\paragraph{Exact Combinatorial Optimization with GCNNs~\cite{GasseCFCL19}.}
This paper serves as one of the backbones of our research; its main focus is to detail how MIPS can potentially be solved much quicker than a
traditional solver by using GNNs (specifically GCNNs). It did this by training its model using imitation learning (using the strong branching expert
rule) and was able to effectively produce outputs for problem instances much greater than what they were trained on.

\paragraph{State of the Art Exact Solver.}
There has been a lot of progress on the symmetric TSP in the last century. With the increase in the number of nodes, there is a super-polynomial (at
least exponential) explosion in the number of potential solutions. This makes the TSP problem difficult to solve on two parameters, the first being
finding a global shortest route as well as reducing the computation complexity in finding this route. \texttt{Concorde}~\cite{concorde}, written in the
ANSI C programming language is widely recognized as the fastest state-of-the-art (SOTA) exact TSP solution for large instances.

\section{Preliminary}
\subsection{Integer Linear Programming Formulation of TSP}
We first formulate TSP in terms of \emph{Integer Linear Programming}. Given an undirected weighted group \(\mathcal{G} = (\mathcal{E}, \mathcal{V})\),
we label the nodes with numbers \(1, \ldots, n\) and define
\[
	x_{ij}\coloneqq \begin{dcases}
		1, & \text{if }(i, j)\in \mathcal{E}^\prime                       \\
		0, & \text{if } (i, j)\in \mathcal{E}\setminus\mathcal{E}^\prime,
	\end{dcases}
\]
where \(\mathcal{E}^\prime\subset \mathcal{E}\) is a variable which can be viewed as a compact representation of all variables \(x_{ij}\), \(\forall i, j\).
Furthermore, we denote the weight on edge \((i, j)\) by \(c_{ij}\), then for a particular TSP problem instance, we can formulate the problem as follows.
\begin{equation}\label{formula:TSP}
	\begin{aligned}
		\min & \sum _{i=1}^{n}\sum _{j\neq i,j=1}^{n}c_{ij}x_{ij} &  &                      \\
		     & \sum _{i=1,i\neq j}^{n}x_{ij}=1                    &  & j=1,\ldots ,n;       \\
		     & \sum _{j=1,j\neq i}^{n}x_{ij}=1                    &  & i=1,\ldots ,n;       \\
		     & u_{i}-u_{j}+nx_{ij}\leq n-1                        &  & 2\leq i\neq j\leq n; \\
		     & 1\leq u_{i}\leq n-1                                &  & 2\leq i\leq n;       \\
		     & x_{ij}\in \{0,1\}                                  &  & i,j=1,\ldots ,n;     \\
		     & u_{i}\in \mathbb{Z}                                &  & i=2,\ldots ,n.
	\end{aligned}
\end{equation}

This is the Miller-Tucker-Zemlin formulation~\cite{MTZ-formulation}. Note that in our case, since we are solving TSP exactly, all variables are
integers. This type of integer linear programming is sometimes known as \emph{pure integer programming}.

\subsection{Solving the Integer Linear Program}
Since integer programming is an NP-Hard problem, there is no known polynomial algorithm that can solve this explicitly. Hence, the modern approach
to such a problem is to \emph{relax} the integrality constraint, which makes \autoref{formula:TSP} becomes continuous linear programming (LP),
whose solution provides a lower bound to \autoref{formula:TSP} since it is a relaxation, and we are trying to find the minimum.

Since an LP is a convex optimization problem, we have many polynomial-time algorithms to solve the relaxed version. After obtaining a relaxed solution, if such LP relaxed
solution respects the integrality constraint, we see that it's indeed a solution to \autoref{formula:TSP}. But if not, we can simply divide the original relaxed LP into two
sub-problems by \emph{splitting the feasible region} according to a variable that does not respect integrality in the current relaxed LP solution \(\bm{x}^\ast\),
\begin{equation}\label{eq:branch-and-bound}
	x_{i} \leq \left\lfloor x_{i}^\ast \right\rfloor\lor x_{i} \geq \left\lceil x_{i}^\ast \right\rceil,\qquad \exists i\leq p\mid x_{i} ^\ast \notin \mathbb{\MakeUppercase{z}}.
\end{equation}

We see that by adding such additional constraints in two sub-problems respectively, we get a recursive algorithm called \emph{Branch-and-Bound}~\cite{B&B.ch7}.
The branch-and-bound algorithm is widely used to solve integer programming problems. We see that the key step in the branch-and-bound algorithm is selecting a non-integer
variable to \underline{branch on} in \autoref{eq:branch-and-bound}. And as one can expect, some choices may reduce the recursive searching tree
significantly~\cite{B&B.branching-impact}, hence the \emph{branching rules} are the core of modern combinatorial optimization solvers, and it has been the focus of extensive
research~\cite{B&B-branching-rules-research-1, B&B-branching-rules-research-2, B&B-branching-rules-research-3, B&B-branching-rules-research-4}.

\subsection{Branching Strategy}
There are several popular strategies~\cite{branching-rules-revisited} used in modern solvers.

\paragraph{Strong branching.}
Strong branching is guaranteed to result in the smallest recursive tree by computing the expected bound improvement for \emph{each} candidate variable before branching by
finding solutions of two LPs for every candidate. However, this is extremely computationally expensive~\cite{Finding-cuts-in-the-TSP}.

\paragraph{Hybrid branching.}
Hybrid branching computes a strong branching score at the beginning of the solving process, but gradually switches to other methods like Conflict
score, Most Infeasible branching, or some other, hand-crafted, combinations of the above~\cite{branching-rules-revisited, B&B-branching-rules-research-4}.

\paragraph{Pseudocost branching.}
This is the default branching strategy used in \texttt{SCIP}. By keeping track of each variable \(x_{i}\) the change in the objective function
when this variable was previously chosen as the variable to branch on, the strategy then chooses the variable that is predicted to have the most
change on the objective function based on past changes when it was chosen as the branching variable~\cite{B&B-branching-rules-research-2}.

\section{Problem Formulation}
In order to solve TSP with ILP efficiently, we use the branch-and-bound algorithm. Specifically, we want to take advantage of the fast inference
time and the learning ability of the model, hence we choose to learn the most powerful branching strategy known: strong branching.
Our objective is then to learn a branching strategy without expensive evaluation. Since this is a discrete-time control process, we model the
problem by Markov Decision Process (MDP)~\cite{howard1960dynamic}.

\subsection{Markov Decision Process (MDP)}
Given a regular Markov decision process \(\mathcal{M} \coloneqq (\mathcal{S}, \mathcal{A}, p_{\mathrm{init}}, p_{\mathrm{trans}}, R)\), we have the
state space \(\mathcal{S}\), action space \(\mathcal{A}\), initial state distribution
\(p_{\mathrm{init}}\colon \mathcal{S} \to \mathbb{R}_{\geq 0}\), state transition distribution
\(p_{\mathrm{trans}}\colon \mathcal{S}\times \mathcal{A}\times \mathcal{S} \to \mathbb{R}_{\geq 0}\) and the reward function
\(R\colon \mathcal{S} \to \mathbb{R}\). One thing to note is that the reward function \(R\) need not be deterministic. In other words, we can
define \(R\) as a random function that will take a value based on a particular state in \(\mathcal{S}\) with some randomness. Note that if \(R\)
in \(\mathcal{M}\) is equipped with any kind of randomness, we can write the reward \(r_t\) at time \(t\) as
\(r_t\sim p_{\mathrm{reward}}(r_t\mid s_{t-1}, a_{t-1}, s_t)\).
This can be converted into an equivalent Markov Decision Process \(\mathcal{M}^\prime\) with a deterministic reward function \(R^\prime\), where
the randomness is integrated into parts of the states.
With an action policy \(\pi \colon \mathcal{A}\times \mathcal{S}\to \mathbb{R}_{\geq 0}\) such that the action \(a_t\) taken at time \(t\) is
determined by \(a_t\sim \pi(a_t\mid s_t)\), we see that an MDP can be unrolled to produce a \emph{trajectory} composed by state-action pairs as
\(\tau = (s_0, a_0, s_1, a_1, \ldots)\) which obeys the joint distribution
\[
	\tau \sim \underbrace{p_{\mathrm{init}}(s_0)}_{\text{initial state}}\prod_{t = 0}^{\infty} \underbrace{\pi(a_t\mid s_t)}_{\text{next action}}\underbrace{p_{\mathrm{trans}}(s_{t+1}\mid a_t, s_t)}_{\text{next state}}
\]

\subsection{Partially Observable Markov Decision Process (PO-MDP)}
Following from the same idea as MDP, the PO-MDP setting deals with the case that when the \emph{complete} information about the current MDP state
\(\mathcal{S}\) is unavailable or not necessarily for decision-making~\cite{ASTROM1965174}.
Instead, in our case, only a partial \emph{observation} \(o\in \Omega\) is available, where \(\Omega\) is called the \emph{partial state space}.
We can use an active perspective to view the above model; namely, we are merely applying an \underline{observation function} \(O\colon \mathcal{S}\to \Omega\)
to the current state \(s_t\) at each time step \(t\). Hence, we define a PO-MDP \(\widetilde{\mathcal{M}}\) as a tuple
\(\widetilde{\mathcal{M}} \coloneqq (\mathcal{S}, \mathcal{A}, p_{\mathrm{init}}, p_{\mathrm{trans}}, R, O)\).
Within this setup, a trajectory of PO-MDP takes form as \(\tau = (o_0, r_0, a_0, o_1, r_1, a_1, \ldots)\), where \(o_t\coloneqq O(s_t)\) and
\(r_t\coloneqq R(s_t)\). It is important to note that here \(r_t\) still depends on the state of the OP-MDP, \emph{not} the observation.
We introduce a convenience variable \(h_t\colon (o_0, r_0, a_0, \ldots, o_t, r_t)\in \mathcal{H}\), which represents the PO-MDP history at time step
\(t\) \emph{without the action} \(a_t\).  Due to the non-Markovian nature of the trajectories,
\(o_{t+1}, r_{t+1}\not\perp h_{t-1} \mid o_t,r_t,a_t\), the decision-maker must take the whole history of observations, rewards and actions into account
to decide on an optimal action at the current time step \(t\). We then see that the action policy for PO-MDP takes the form
\(\widetilde{\pi}\colon \mathcal{A}\times \mathcal{H}\to \mathbb{R}_{\geq 0}\) such that \(a_t\sim \pi (a_t\mid h_t)\).

\subsection{Markov Control Problem}
We define the MDP control problem as that of finding a policy \(\pi ^\ast\colon \mathcal{A}\times \mathcal{S}\to \mathbb{R}_{\geq 0}\)
which is optimal with respect to the expected total reward. That is,
\[
	\pi^\ast = \underset{\pi}{\arg\max} \lim_{T\to \infty}\mathbb{E}_\tau\left[\sum_{t=0}^T r_t\right]
\]
where \(r_t\coloneqq R(s_t)\). To generalize this into a PO-MDP control problem, similar to the MDP control problem, the objective is to find a policy
\(\widetilde{\pi}^\ast\colon \mathcal{A}\times \mathcal{H}\to \mathbb{R}_{\geq 0}\) such that it maximizes the expected total rewards.
By slightly abusing the notation, we simply denote this learned policy by \(\widetilde{\pi}^\ast\) where the objective function is completely the same as in the MDP case.

\section{Methodology}
Since the branch-and-bound variable selection problem can be naturally formulated as a Markov decision process, a natural machine learning algorithm
to use is reinforcement learning~\cite{sutton2018reinforcement}. Specifically, since there are some SOTA integers programming solvers out there,
\texttt{Gurobi}~\cite{gurobi}, \texttt{SCIP}~\cite{SCIP}, etc., we decided to try imitation learning~\cite{Imitation-Learning-A-Survey-of-Learning-Methods}
by learning directly from an expert branching rule. There are some related works in this approach~\cite{GasseCFCL19} aiming to tackle
\emph{mixed integer linear programming} (MILP) where only a portion of variables have integral constraints, while other variables can be real numbers.
Our approach extends this further. We are focusing on TSP, which not only is pure integer programming, but also the variables can only take values from
\(\{0, 1\}\).

\subsection{Learning Pipeline}
Our learning pipeline is as follows: we first create some random TSP instances and turn them into ILP.
Then, we use imitation learning to learn how to choose the \emph{branching target} at each branching.
Our GNN model produces a set of actions with the probability corresponding to each possible action (in our case, which variable to branch). We then
use \emph{Cross-Entropy Loss} to compare our prediction to the result produced by \texttt{SCIP} and complete one iteration.

\begin{figure}
	\centering
	\def\svgwidth{\columnwidth}
	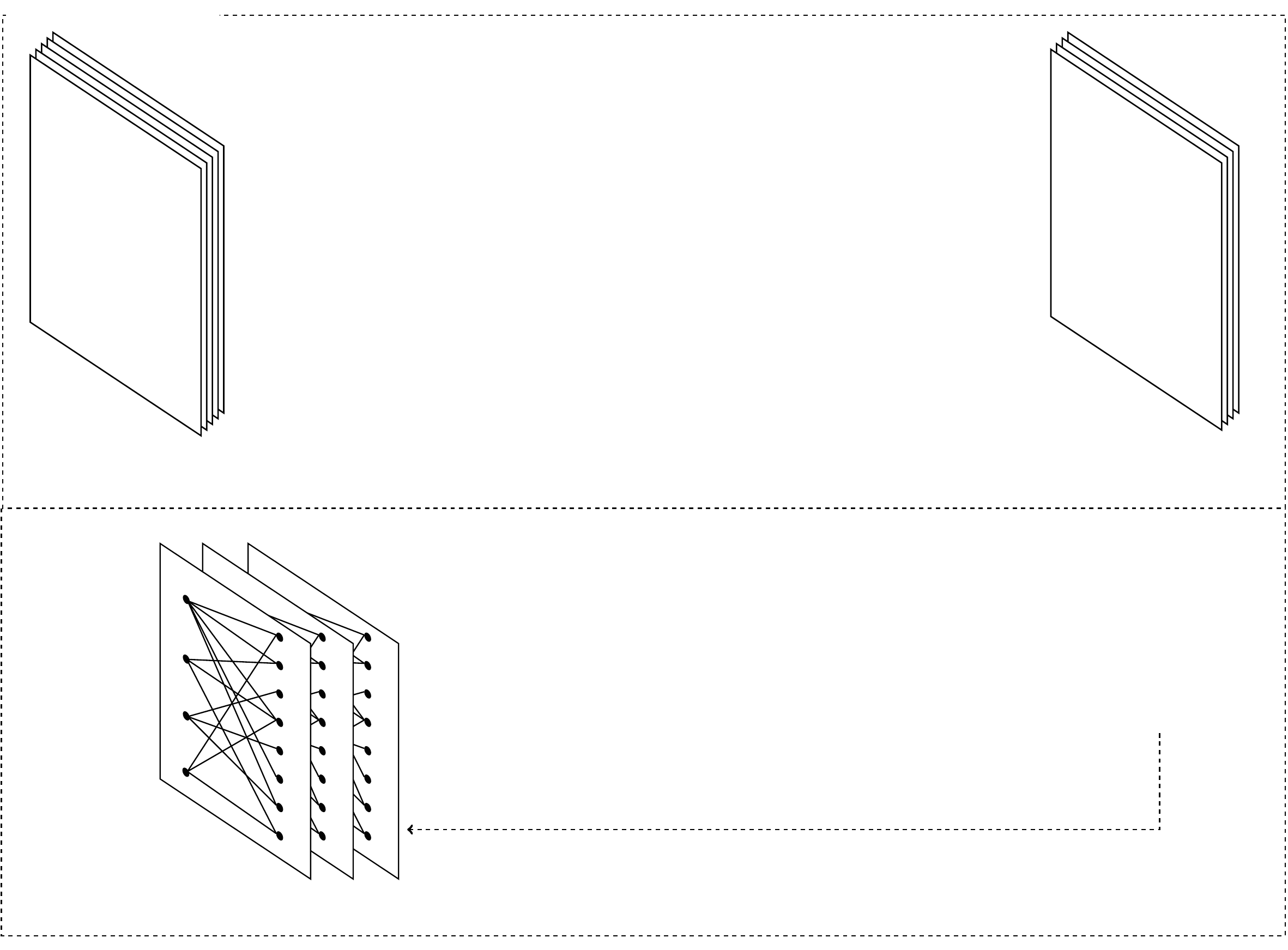

\end{figure}

\paragraph{Instances Generation.}
For each TSP instance, we randomly generate the coordinates for every node and formulate it by using
Miller-Tucker-Zemlin formulation~\cite{MTZ-formulation} and record it in the linear programming format called \texttt{instances\_*.lp}
via \texttt{CPlex}~\cite{cplex2009v12}.

\paragraph{Samples Generation.}
By passing every \texttt{instances\_*.lp} to \texttt{SCIP}, we can record the branching decision solver made when solving it.
The modern solver usually uses a mixed branching strategy to balance the running time, but since we want to learn the best branching strategy, we ask
\texttt{SCIP} to use a strong branch with some probability when branching, and only record the state and branching decision (state-action pairs)
\(\mathcal{\MakeUppercase{d}} = \left\{(s_{i} , \bm{a} _{i} ^\ast)\right\}_{i = 1}^N\) when \texttt{SCIP} uses strong branch.

\paragraph{Imitating Learning.}
We learn our policy \(\widetilde{\pi} ^\ast\) by minimizing the cross-entropy loss
\[
	\mathcal{\MakeUppercase{l}} (\theta ) = - \frac{1}{N}\sum\limits_{(\bm{s}, \bm{a}^\ast)\in \mathcal{\MakeUppercase{d}} }\log \widetilde{\pi}_\theta (\bm{a} ^\ast \mid \bm{s} )
\]
to train by behavioral cloning~\cite{Efficient-Training-of-artificial-Neural-Networks-for-Autonomous-Navigation} from the state-action pairs we
recorded.

\paragraph{Evaluation.}
We evaluate our model on TSP instances with various sizes to see the generalization ability.
To compare the result of default \texttt{SCIP} performance to our learned branching strategy, we look at the wall-time needed for solving.
Also, we look at the performance of the SOTA TSP solver to see the performance between our naive formulation and solving strategy and the SOTA solver
which fully exploits the problem structure of TSP.

\subsection{Policy Parametrization by GCNN}
We use GCNN~\cite{Gori2005ANM, https://doi.org/10.48550/arxiv.1312.6203, paschos2014applications} to parametrize
the variable selection policy. This specific choice is due to the natural problem structure of the branch and bound
decision process since we equipped our input with a \emph{bipartite graph}~\cite{GasseCFCL19}, and utilize the message passing mechanism
inherited by GCNN. Other models are compared in the Gasse et al.'~\cite{GasseCFCL19}, and GCNN outperforms all other models like LMART, SVMRANK,
TREES, etc.

\section{Experiments}
Our implementation of the imitating learning model generally follows the work by Gasse et al~\cite{GasseCFCL19} and depends on several
packages~\cite{gurobi, SCIP, cplex2009v12,prouvost2020ecole}. We test the generalization ability of our model trained with TSP10 and TSP15 on TSP
instances with various sizes using GreatLakes with one A100 GPU and 8GB, 16 cores CPU. The figures below plot the wall-time needed for our model to
solve a particular TSP instance as a direct comparison to our baseline \texttt{SCIP}, the solver we're imitating during the training phase, and also
compare to the SOTA TSP solver \texttt{Concorde}.

\begin{figure}[H]
	\begin{minipage}[b]{0.5\linewidth}
		\centering
		\includegraphics[width=\linewidth]{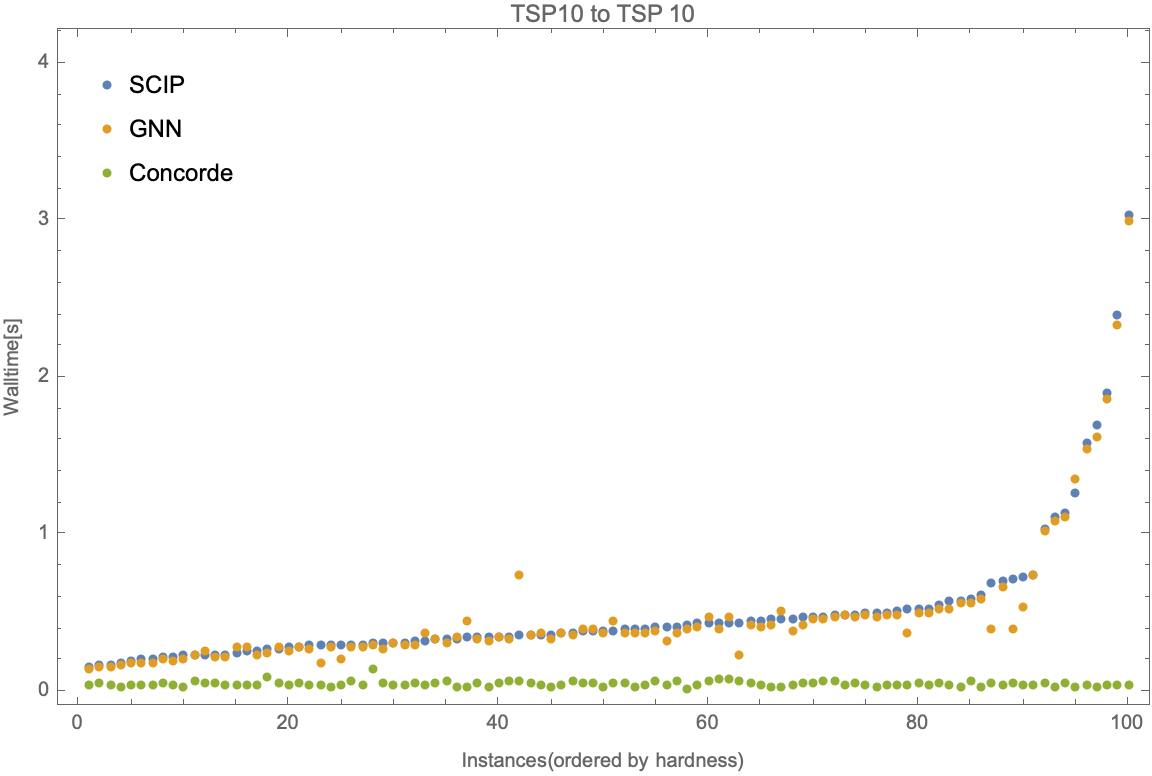}
	\end{minipage}
	\begin{minipage}[b]{0.5\linewidth}
		\centering
		\includegraphics[width=\linewidth]{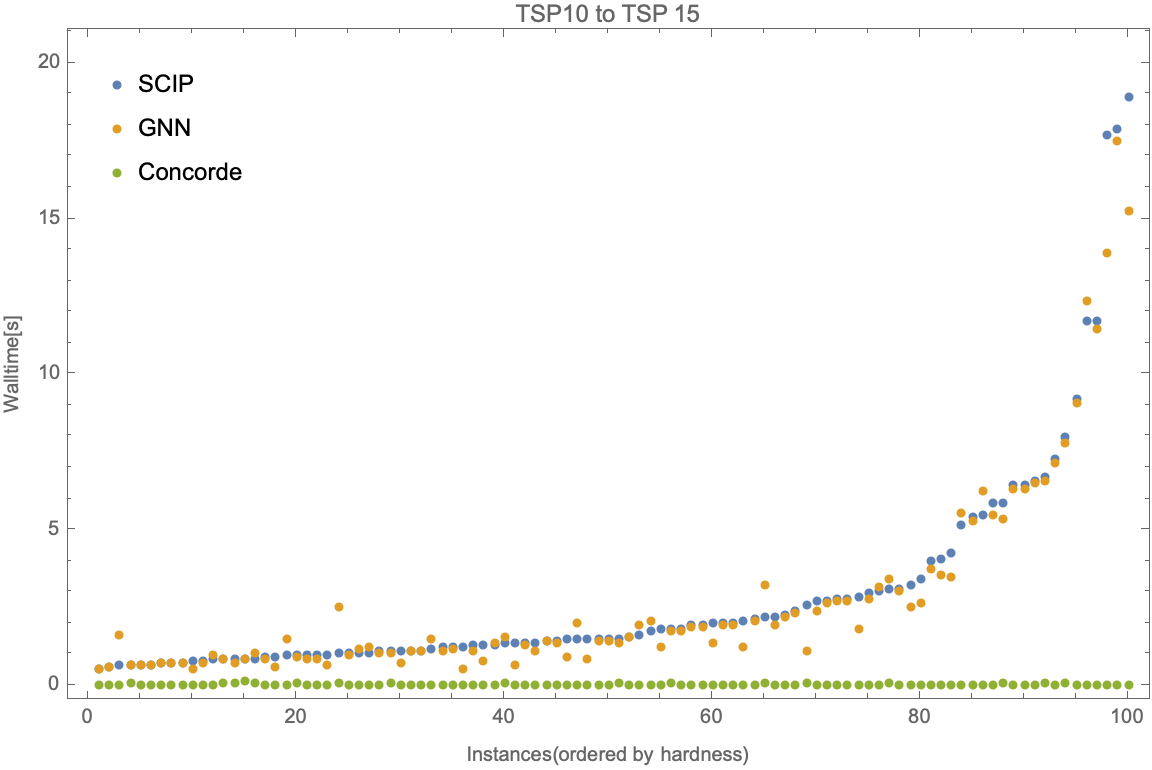}
	\end{minipage}
	\begin{minipage}[b]{0.5\linewidth}
		\centering
		\includegraphics[width=\linewidth]{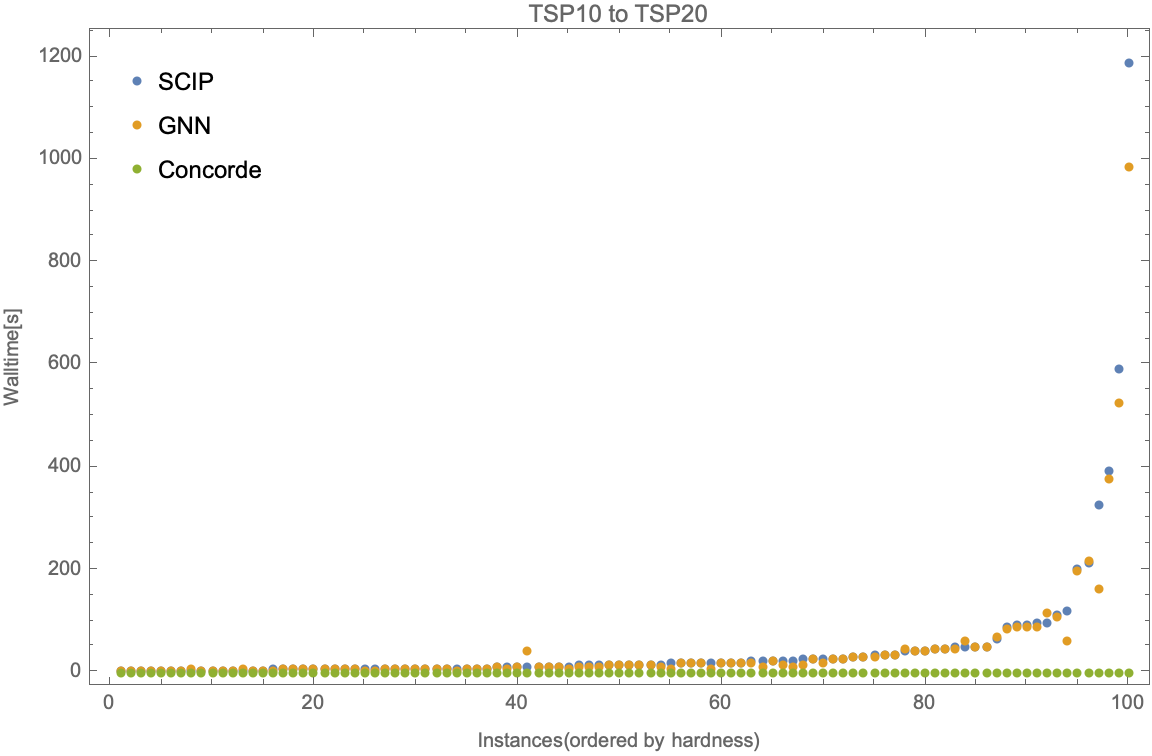}
	\end{minipage}
	\begin{minipage}[b]{0.5\linewidth}
		\centering
		\includegraphics[width=\linewidth]{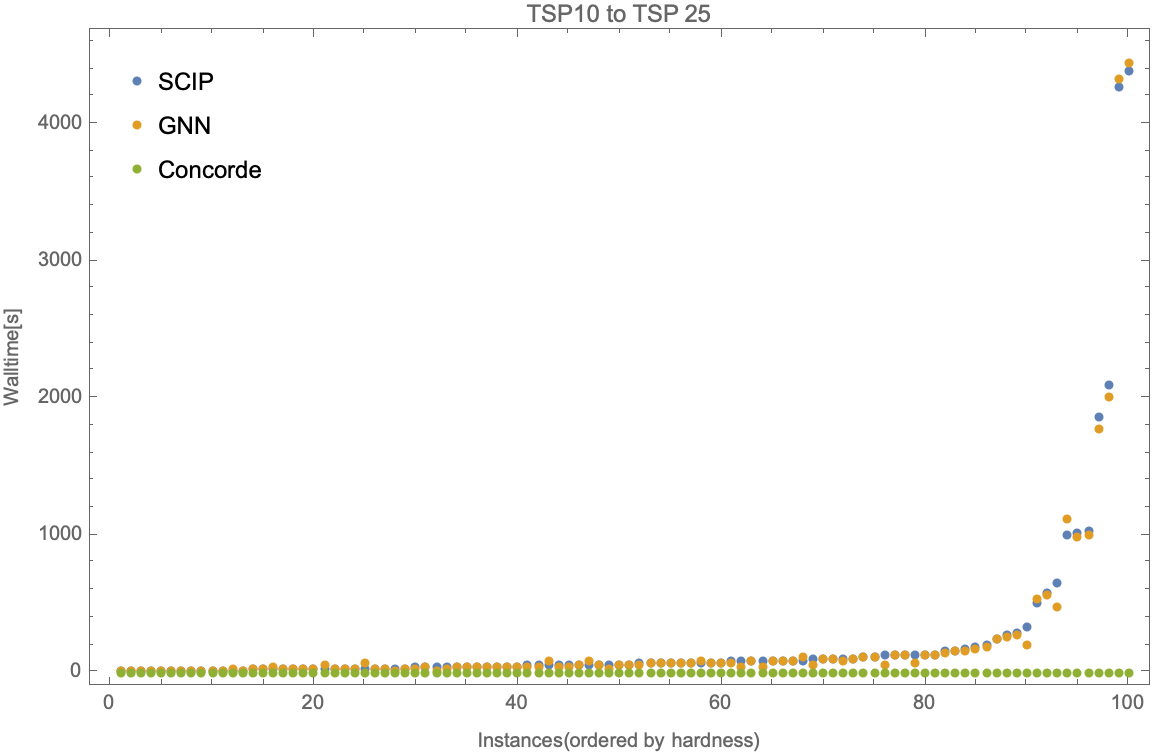}
	\end{minipage}
	\caption{Result of model trained on TSP10 generalizes to TSP with various sizes.}
	\label{fig:tsp10}
\end{figure}

\begin{figure}[H]
	\begin{minipage}[b]{0.5\linewidth}
		\centering
		\includegraphics[width=\linewidth]{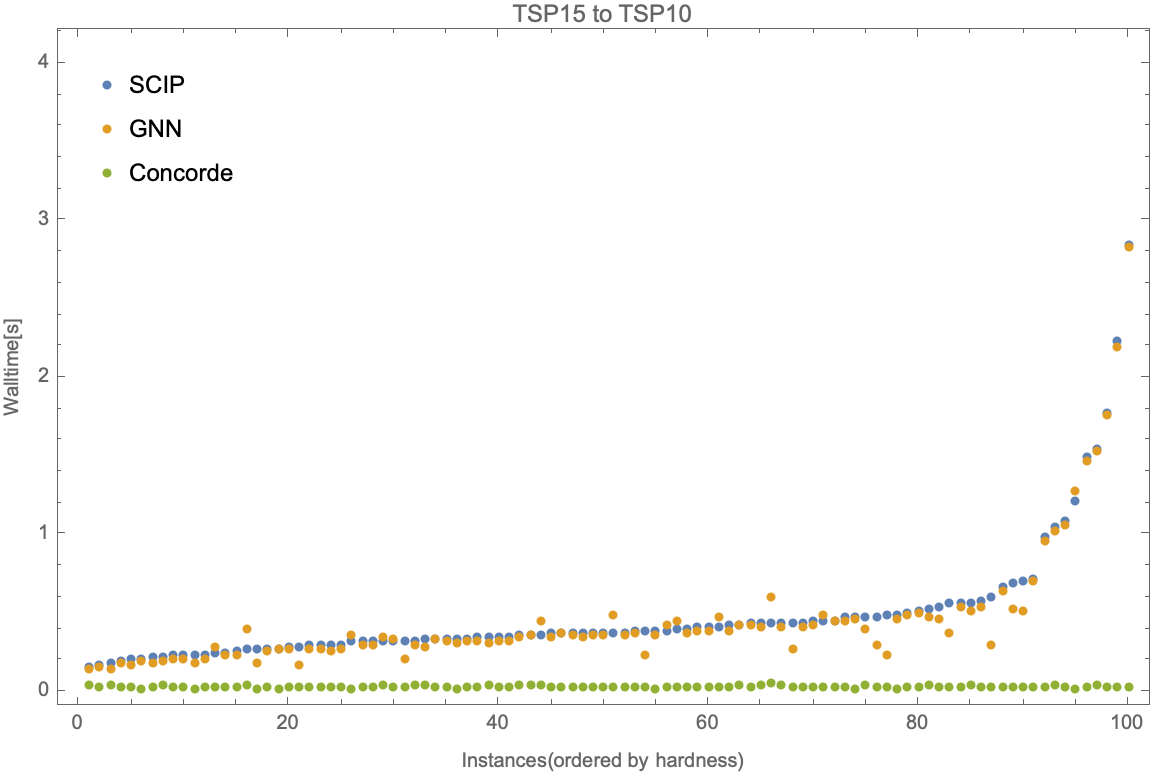}
	\end{minipage}
	\begin{minipage}[b]{0.5\linewidth}
		\centering
		\includegraphics[width=\linewidth]{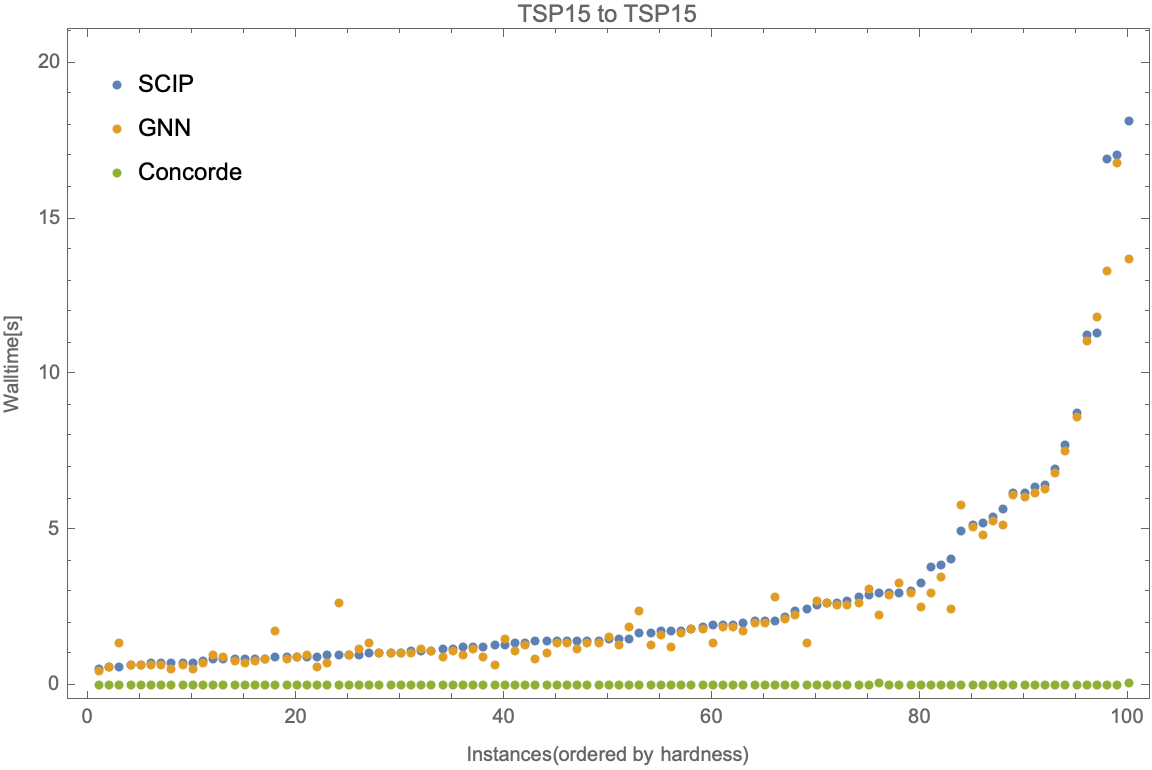}
	\end{minipage}
	\begin{minipage}[b]{0.5\linewidth}
		\centering
		\includegraphics[width=\linewidth]{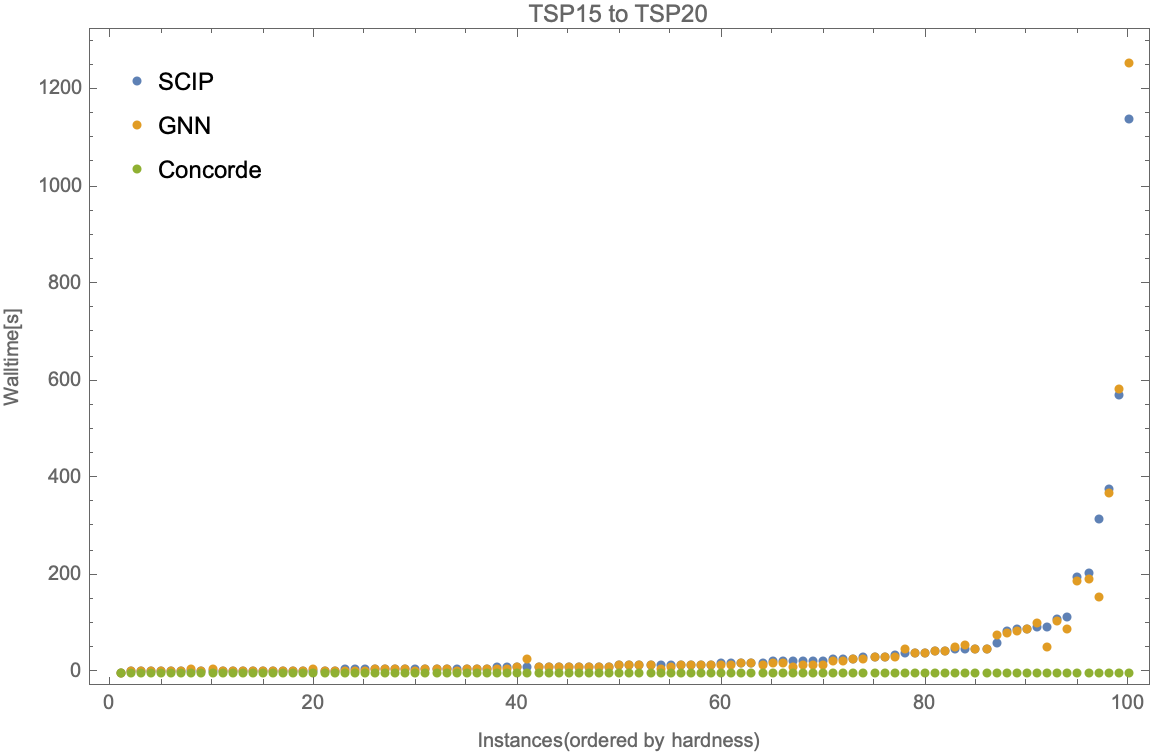}
	\end{minipage}
	\begin{minipage}[b]{0.5\linewidth}
		\centering
		\includegraphics[width=\linewidth]{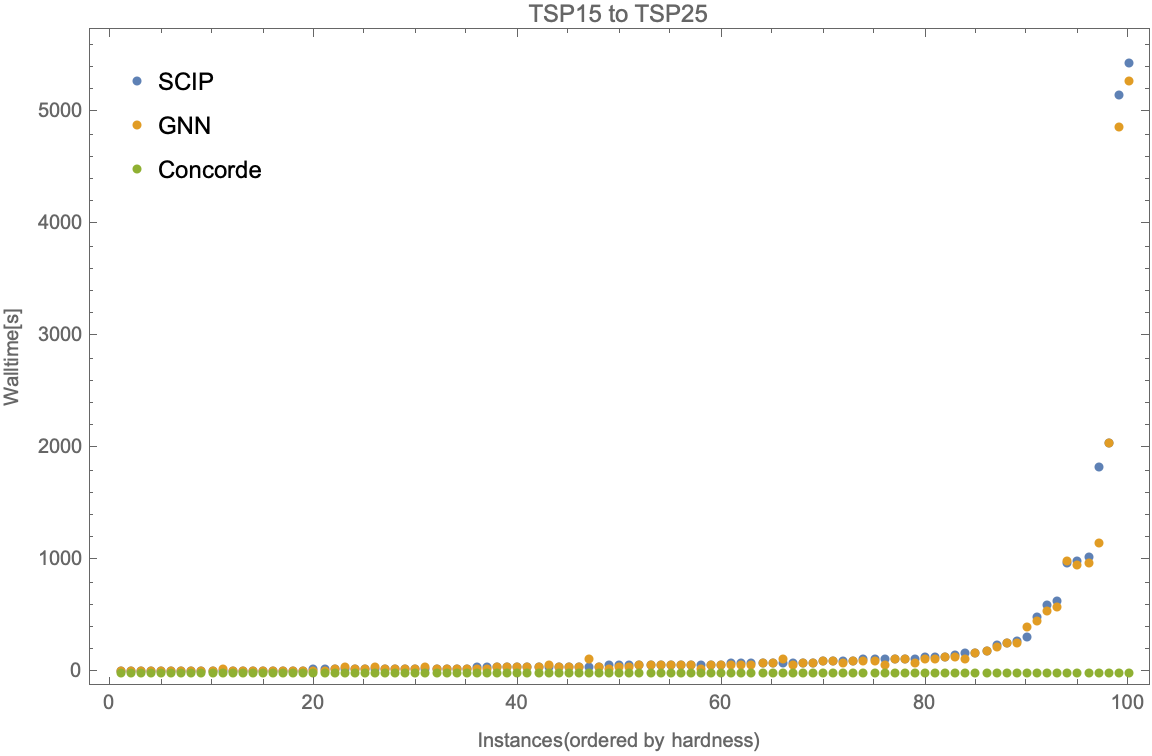}
	\end{minipage}
	\caption{Result of model trained on TSP15 generalizes to TSP with various sizes.}
	\label{fig:tsp15}
\end{figure}

\autoref{fig:tsp10} and \autoref{fig:tsp15} show the testing result of the models trained on TSP10 and TSP15, respectively. The analytical result
is also shown in \autoref{table:tsp10} and \autoref{table:tsp15}. Note that since some instances are much harder than others, we divide the data by
the wall-time needed for \texttt{SCIP} and do a detailed comparison. Also, we compare the performance between \texttt{Concorde} and the TSP solving
API provided by \texttt{Gurobi} in \autoref{fig:tsp50} and \autoref{fig:tsp100}. The result is similar when the TSP size is small, so
we didn't include the \texttt{Gurobi} result in the following plots.

\begin{table}[H]
	\centering
	\begin{tabular}{c|cccccccc}
		\toprule
		\multirow{2}*{Test Size} & \multicolumn{2}{c@{}}{Avg. Walltime(s)} & \multicolumn{3}{c@{}}{Avg. Improvement(s)} & \multicolumn{3}{c@{}}{Avg. Improvement(\%)}                                                                               \\
		\cmidrule(l){2-3}
		\cmidrule(l){4-6}
		\cmidrule(l){7-9}
		                         & \texttt{SCIP}                           & GCNN                                       & All                                         & First 80        & Last 20          & All         & First 80   & Last 20     \\
		\midrule
		TSP10                    & \(0.507\si{s}\)                         & \(0.484\si{s}\)                            & \(0.022\si{s}\)                             & \(0.012\si{s}\) & \(0.063\si{s}\)  & \(4.40\%\)  & \(3.41\%\) & \(5.68\%\)  \\
		\midrule
		TSP15                    & \(2.932\si{s}\)                         & \(2.764\si{s}\)                            & \(0.168\si{s}\)                             & \(0.090\si{s}\) & \(0.481\si{s}\)  & \(5.73\%\)  & \(5.77\%\) & \(5.71\%\)  \\
		\midrule
		TSP20                    & \(50.794\si{s}\)                        & \(44.972\si{s}\)                           & \(5.822\si{s}\)                             & \(0.985\si{s}\) & \(25.174\si{s}\) & \(11.46\%\) & \(7.14\%\) & \(12.66\%\) \\
		\midrule
		TSP25                    & \(238.699\si{s}\)                       & \(231.872\si{s}\)                          & \(6.827\si{s}\)                             & \(3.527\si{s}\) & \(20.028\si{s}\) & \(2.86\%\)  & \(6.52\%\) & \(2.05\%\)  \\
		\bottomrule
	\end{tabular}
	\caption{Model Trained on TSP10}
	\label{table:tsp10}
\end{table}

\begin{figure}[H]
	\begin{minipage}[b]{0.5\linewidth}
		\centering
		\includegraphics[width=\linewidth]{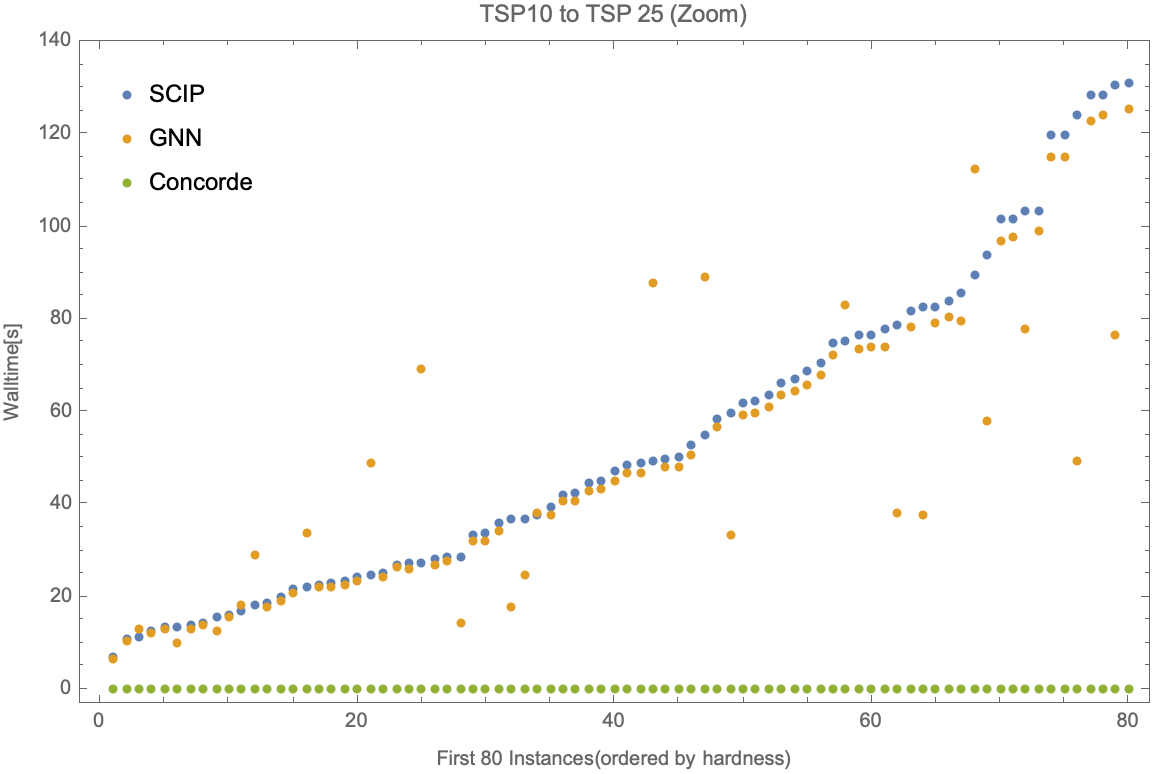}
	\end{minipage}
	\begin{minipage}[b]{0.5\linewidth}
		\centering
		\includegraphics[width=\linewidth]{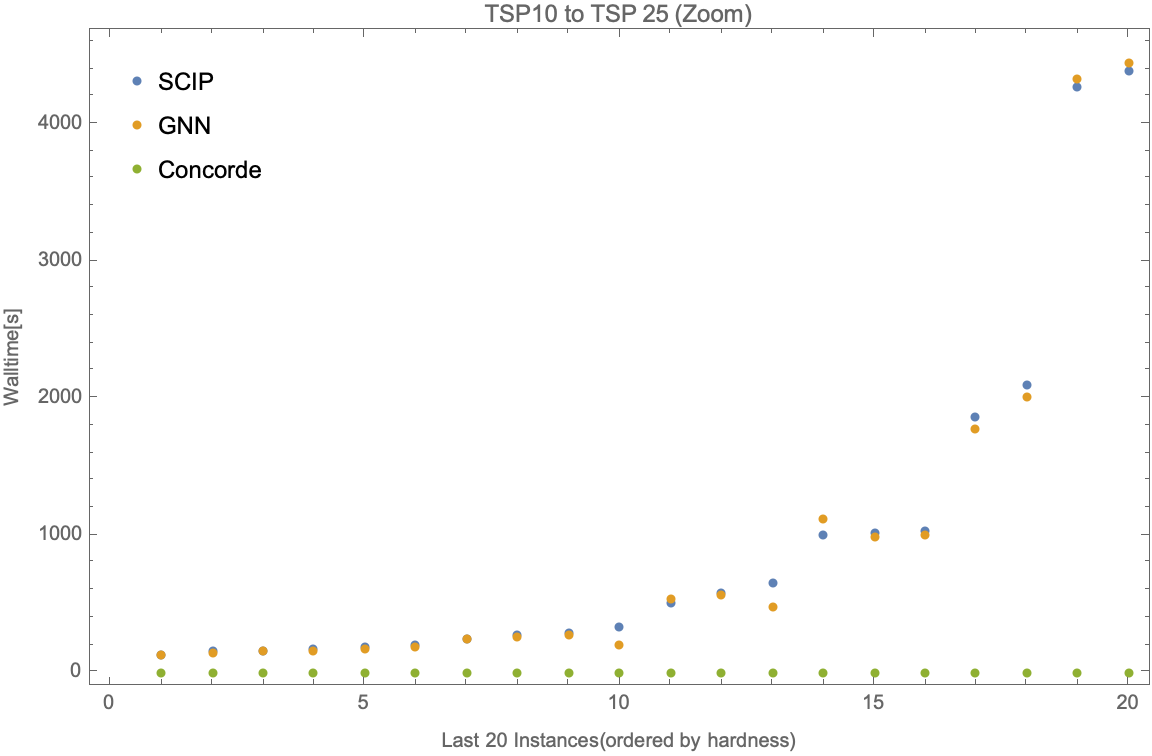}
	\end{minipage}
	\caption{Result of model trained on TSP10 generalizes to TSP25 with zoomed-in.}
	\label{fig:tsp10-zoom}
\end{figure}

\begin{table}[H]
	\centering
	\begin{tabular}{c|cccccccc}
		\toprule
		\multirow{2}*{Test Size} & \multicolumn{2}{c@{}}{Avg. Walltime(s)} & \multicolumn{3}{c@{}}{Avg. Improvement(s)} & \multicolumn{3}{c@{}}{Avg. Improvement(\%)}                                                                              \\
		\cmidrule(l){2-3}
		\cmidrule(l){4-6}
		\cmidrule(l){7-9}
		                         & \texttt{SCIP}                           & GCNN                                       & All                                         & First 80        & Last 20          & All        & First 80   & Last 20     \\
		\midrule
		TSP10                    & \(0.490\si{s}\)                         & \(0.461\si{s}\)                            & \(0.028\si{s}\)                             & \(0.020\si{s}\) & \(0.063\si{s}\)  & \(5.80\%\) & \(5.60\%\) & \(6.07\%\)  \\
		\midrule
		TSP15                    & \(2.822\si{s}\)                         & \(2.661\si{s}\)                            & \(0.161\si{s}\)                             & \(0.050\si{s}\) & \(0.605\si{s}\)  & \(5.70\%\) & \(3.31\%\) & \(7.48\%\)  \\
		\midrule
		TSP20                    & \(49.020\si{s}\)                        & \(47.181\si{s}\)                           & \(1.8389\si{s}\)                            & \(0.878\si{s}\) & \(5.683\si{s}\)  & \(3.75\%\) & \(6.58\%\) & \(2.96\%\)  \\
		\midrule
		TSP25                    & \(256.253\si{s}\)                       & \(239.864\si{s}\)                          & \(16.389\si{s}\)                            & \(3.515\si{s}\) & \(67.886\si{s}\) & \(6.40\%\) & \(6.56\%\) & \( 6.36\%\) \\
		\bottomrule
	\end{tabular}
	\caption{Model Trained on TSP15}
	\label{table:tsp15}
\end{table}

\begin{figure}[H]
	\begin{minipage}[b]{0.5\linewidth}
		\centering
		\includegraphics[width=\linewidth]{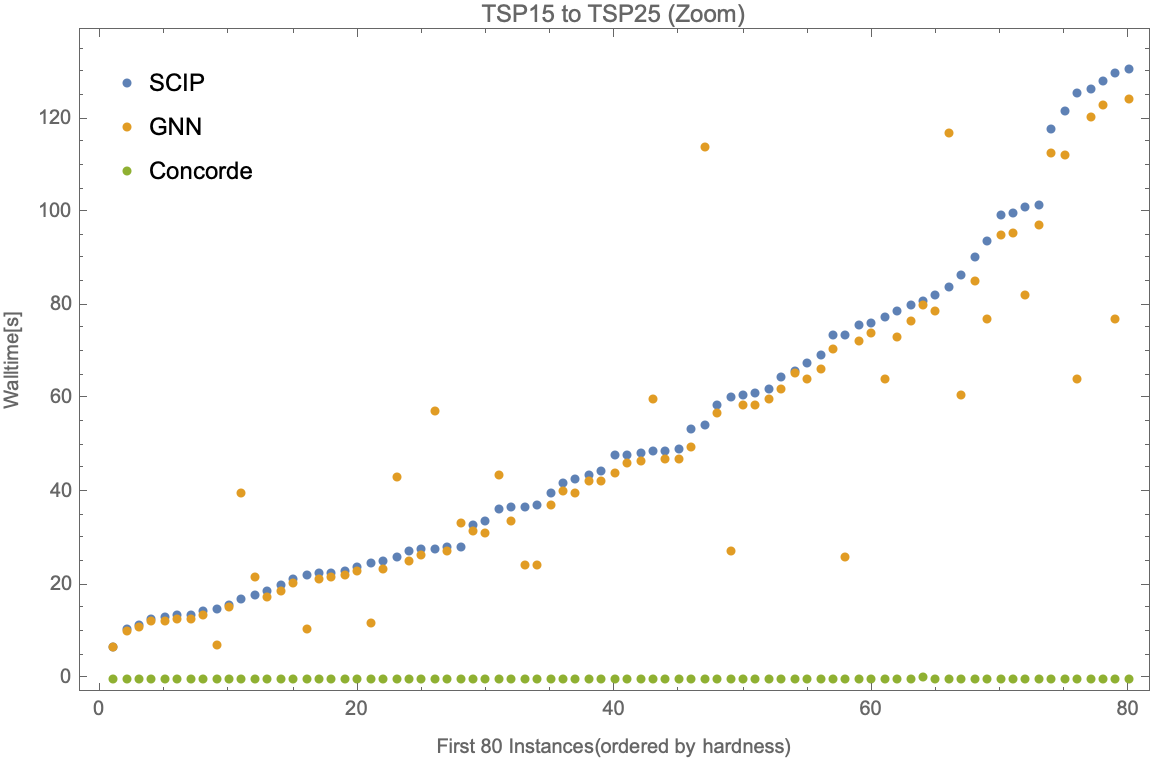}
	\end{minipage}
	\begin{minipage}[b]{0.5\linewidth}
		\centering
		\includegraphics[width=\linewidth]{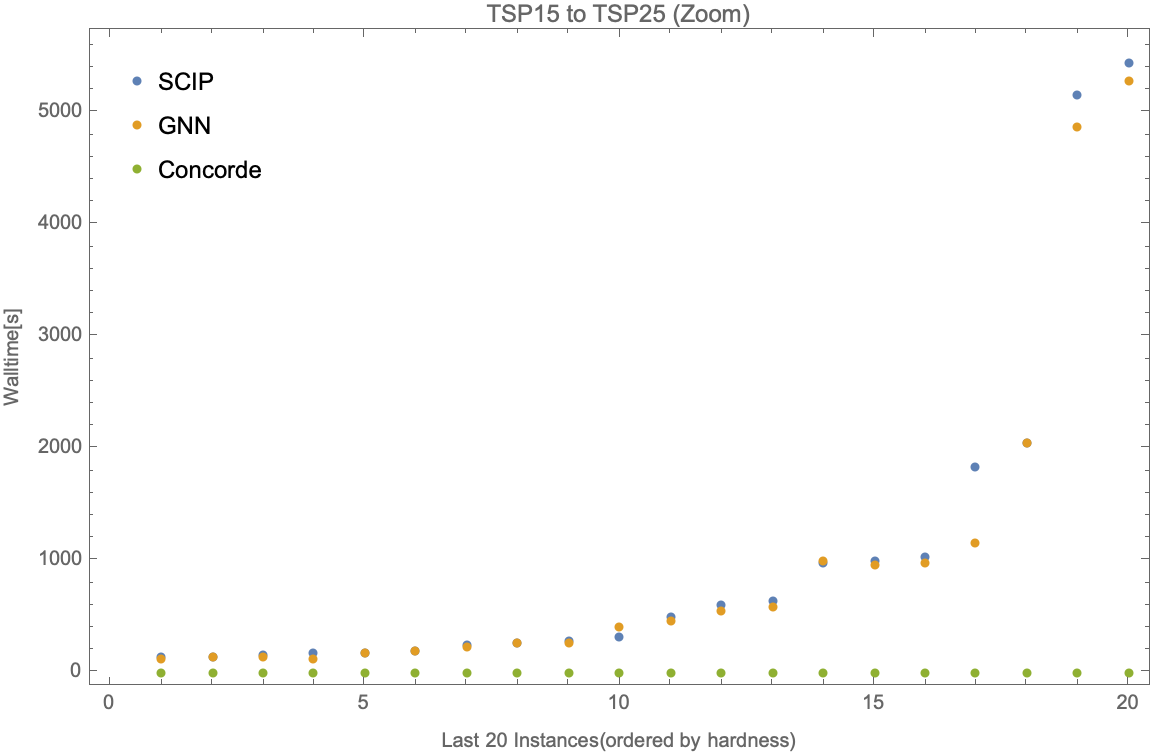}
	\end{minipage}
	\caption{Result of model trained on TSP15 generalizes to TSP25 with zoomed-in.}
	\label{fig:tsp15-zoom}
\end{figure}

The zoomed-in plots for other cases can be found in \autoref{apx:tsp10} and \autoref{apx:tsp15}.

\section{Discussion}
\subsection{Generalization Ability}
We observe that our TSP10 and TSP15 imitation models outperform the \texttt{SCIP} solver on baseline test instances, and
\emph{successfully generalizes to TSP15, TSP20, and TSP25}. They perform significantly better on average than \texttt{SCIP} in difficult-to-solve TSPs as
compared to easier instances. They also perform better in cases of larger test instances like TSP20 and TSP25 as compared to TSP10 and TSP15.
This might be due to an inherent subset structure between TSP10 and TSP20 instances, and similarly TSP15 and TSP25 instances which might not be the
case for smaller test sizes. Unlike other problems, when we formulate TSP as an ILP, the problem size is growing quadratically.\footnote{ This due to the growth rate of edges is quadratic and the number of variables (also constraints) depending on the number of edges directly.}
In other words, when we look at the model performance, the generalization ability from TSP10 to TSP25 is not a \(2.5\times\), but rather a \(6\times\)
generalization in our formulation. By adopting this methodology on a more sophisticated algorithm that formulates TSP linearly, the generalization
ability should remain and the performance will be even better in terms of TSP sizes.

Recent works on finding sub-optimal solutions of TSPs, have not been able to generalize well to large test instances~\cite{learningTSP}. Generalization
ability is one of the most significant properties of Combinatorial Optimization algorithms due to the increasing computational complexity when the problem
size scales up. Finding a sub-optimal solution may be undesirable in a lot of real-world applications since there is no guarantee on the approximation ratio
of all machine learning approaches. Hence, our work is a vital step in this direction.

\subsection{Bottlenecks and Future Work}
There is a huge performance difference between our proposed model (also \texttt{SCIP}) and the SOTA TSP solver, \texttt{Concorde}. Since the proposed
model's backbone is the branch and bound algorithm, by formulating TSP into an ILP, we lost some useful problem structures which can be further exploited by
algorithms used in \texttt{Concorde}. But the existence of a similar pattern of growth in solving time for more difficult instances of larger TSP sizes
even for \texttt{Gurobi} and \texttt{Concorde} is promising (see \autoref{apx:Concorde_Gurobi}), as our imitation model applied to these solvers should
lead to similar time improvements. A major bottleneck is that SOTA solvers like \texttt{Gurobi}, or \texttt{Concorde}, are often licensed, hence not
open-sourced~\cite{gurobi,concorde}. This results in the difficulty of utilizing a stronger baseline and learning from which to get further improvement.

On the other hand, the imitation method can be readily adapted to other algorithms where sequential decision-making is part of the optimization process.
One promising avenue would be a direct adaption to cutting plane methods.\footnote{Specifically, Ecole~\cite{prouvost2020ecole} is working on this. See \url{https://github.com/ds4dm/ecole/issues/319}.}
However, this might be difficult as modern solvers usually utilize different techniques \emph{interchangeably}, which makes a direct adaption non-trivial.

Another concern is that the hyperparameters are not being cross-validated. This is essentially due to computational and hardware limitations.
We can increase the \emph{entropy reward} when calculating cross-entropy for instances, which will motivate the model to be more active when
searching for the optimum on the loss surface. We can also let \texttt{SCIP} use a strong branch with different probabilities, the converging rate may
change and can affect the performance as well.

\section{Conclusion}
Finding exact solutions to combinatorial optimization problems as fast as possible is a challenging avenue in modern theoretical CS. Our proposed method
is a step toward this goal via machine learning. For nearly all exact optimization solving algorithms, there is some kind of \emph{exhaustion} going on
which usually involves decision-making when executing the algorithm. For example, the cutting plane algorithm~\cite{cuttingplane, MARCHAND2002397} also
involves decision-making on variables when it needs to choose a variable to cut. We see that by using our model to replace several such algorithms,
we can speed up the inference time while still retaining a high-quality decision strategy. Furthermore, our experimental results show that the model can
effectively learn such strategies while using less time when inference, which is a promising strategy when applied to other such algorithms.

\section*{Acknowledgement}
I thank Jonathan Moore, Yi Zhou, Shubham Kumar Pandey, and Anuraag Ramesh for conducting the experiment and having insightful discussions.

\newpage
\bibliographystyle{plainnat}
\bibliography{ref}

\newpage
\appendix
\appendixpage

\section{Additional Experimental Results}
\subsection{Model Trained on TSP10}\label{apx:tsp10}
\subsubsection{Full Size Plots}
We include the full-size plots for the testing result on the model trained with TSP10.
\begin{figure}[H]
	\centering
	\includegraphics[width=0.8\linewidth]{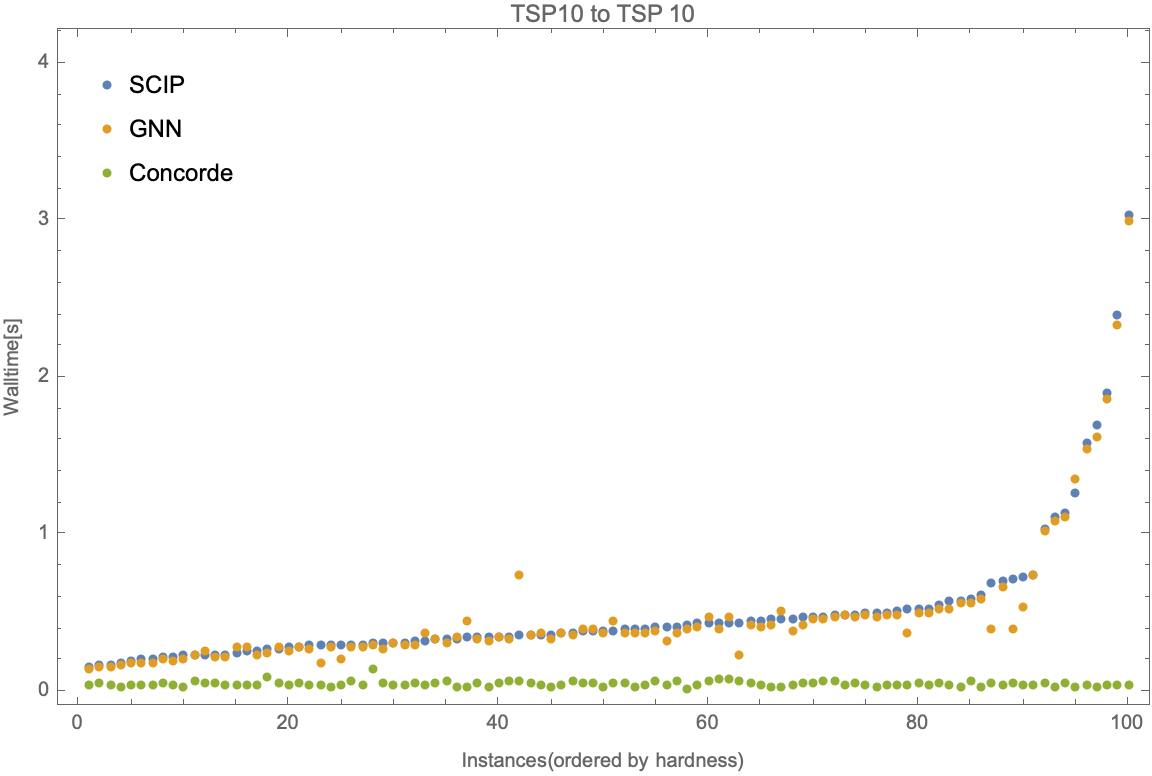}
	\caption{Test on TSP10}
\end{figure}
\begin{figure}[H]
	\centering
	\includegraphics[width=0.8\linewidth]{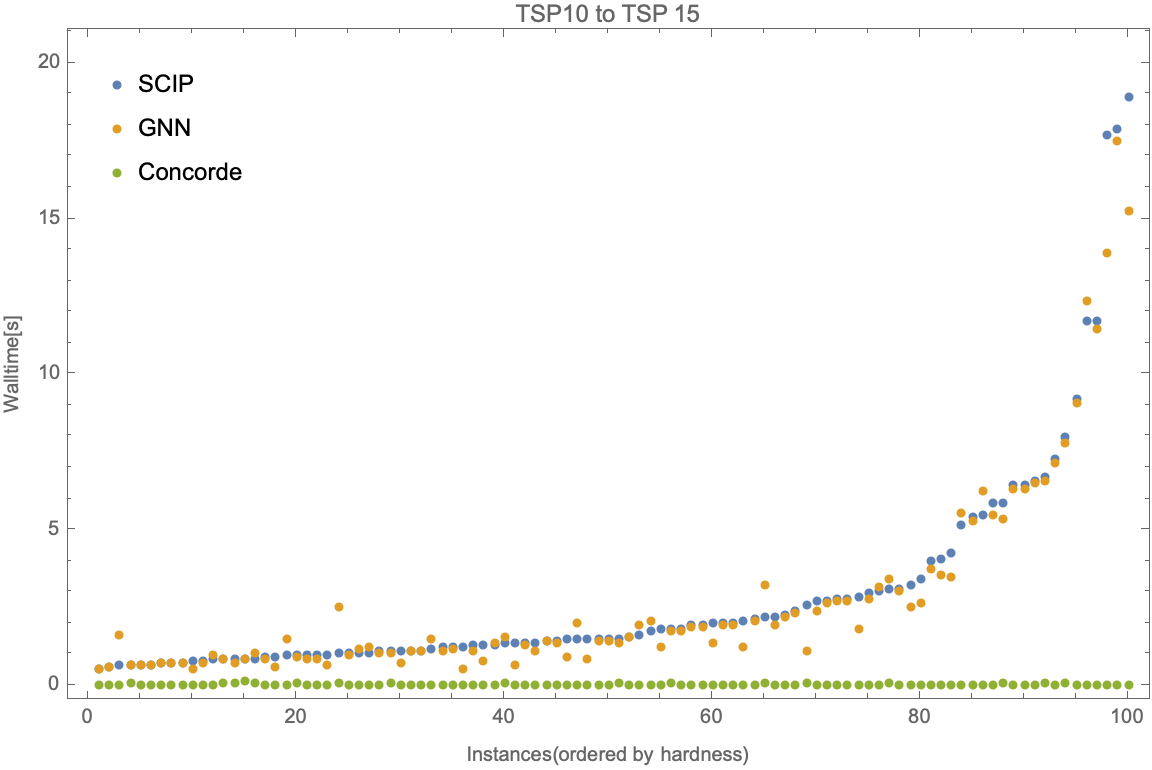}
	\caption{Test on TSP15}
\end{figure}
\begin{figure}[H]
	\centering
	\includegraphics[width=0.8\linewidth]{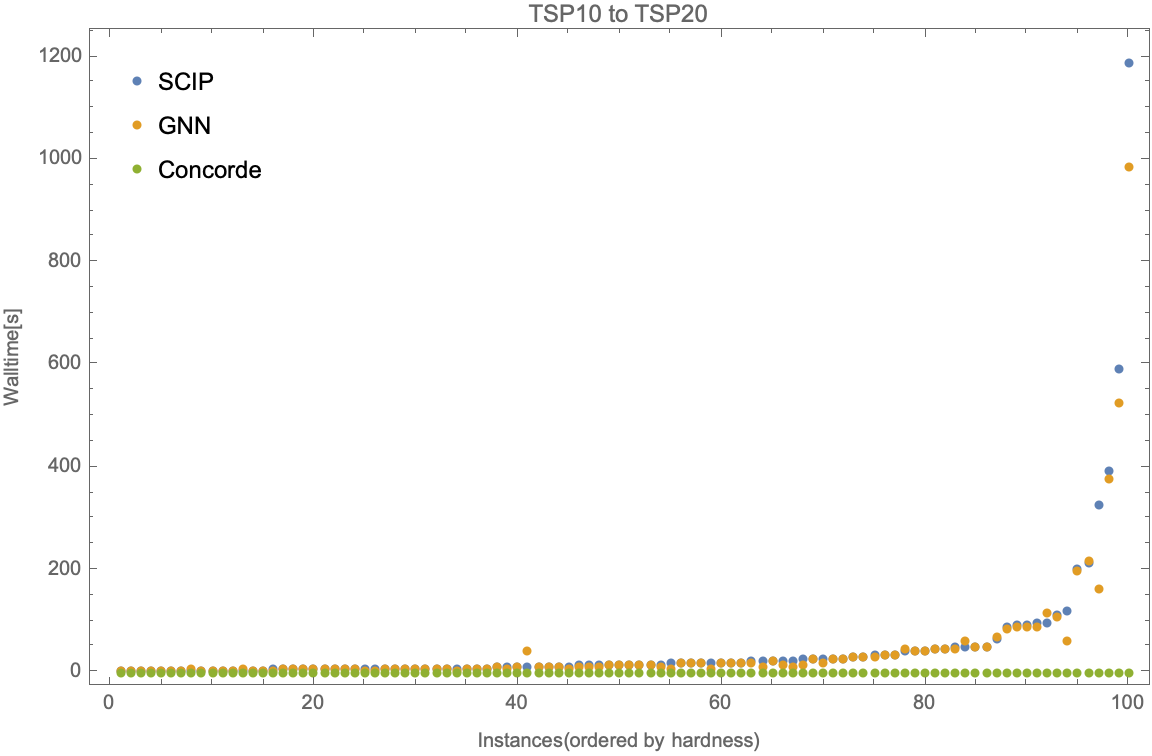}
	\caption{Test on TSP20}
\end{figure}
\begin{figure}[H]
	\centering
	\includegraphics[width=0.8\linewidth]{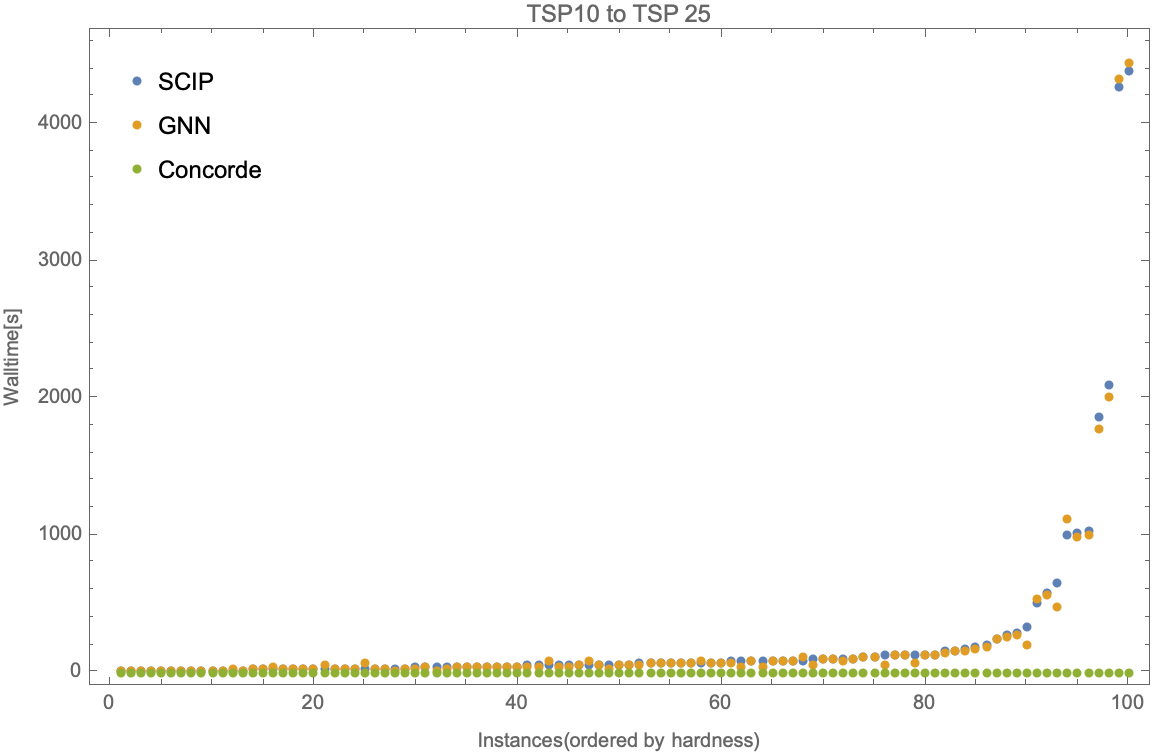}
	\caption{Test on TSP25}
\end{figure}

\subsubsection{Zoom In Plots}
We include the zoom-in plots for the testing result on the model trained with TSP10.
\begin{figure}[H]
	\begin{minipage}[b]{0.5\linewidth}
		\centering
		\includegraphics[width=\linewidth]{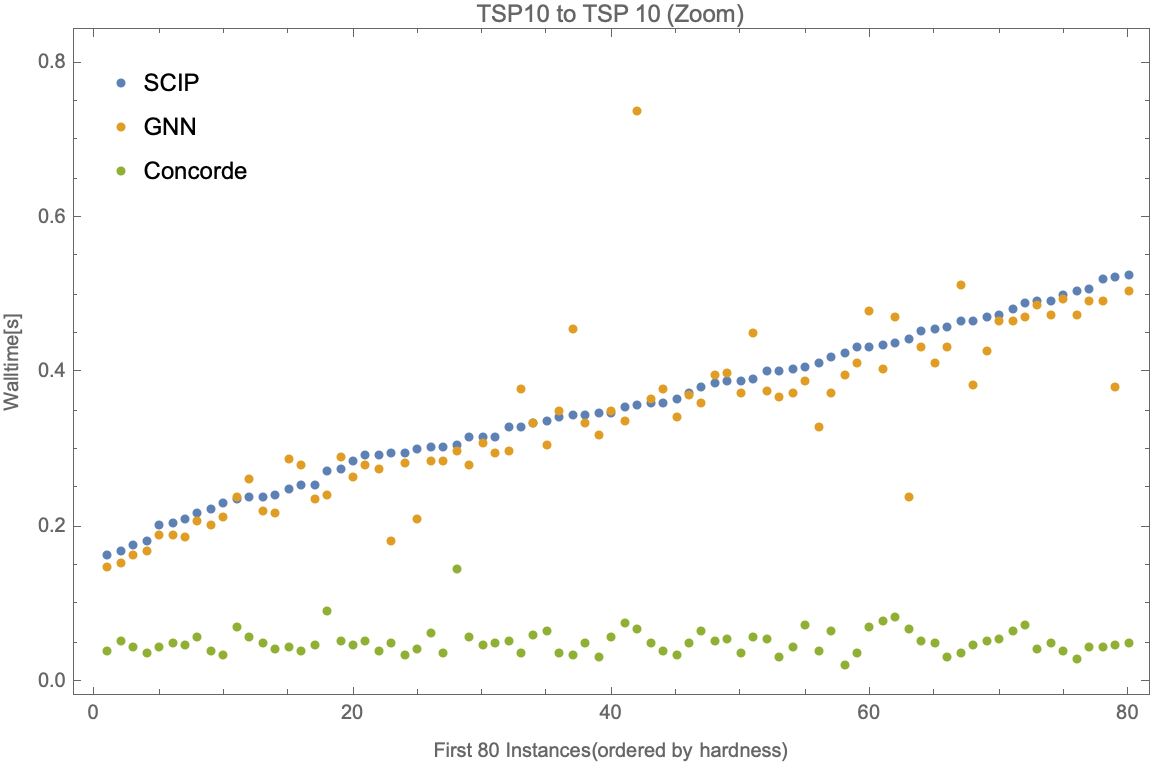}
	\end{minipage}
	\begin{minipage}[b]{0.5\linewidth}
		\centering
		\includegraphics[width=\linewidth]{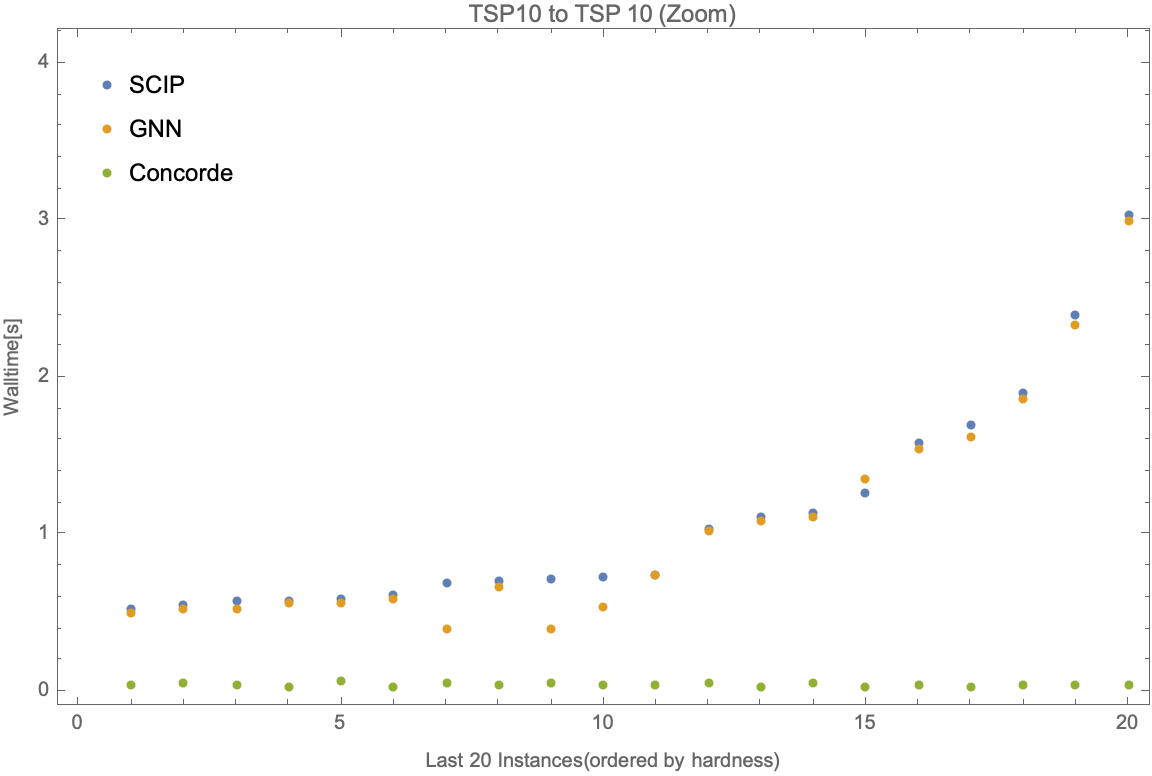}
	\end{minipage}
	\begin{minipage}[b]{0.5\linewidth}
		\centering
		\includegraphics[width=\linewidth]{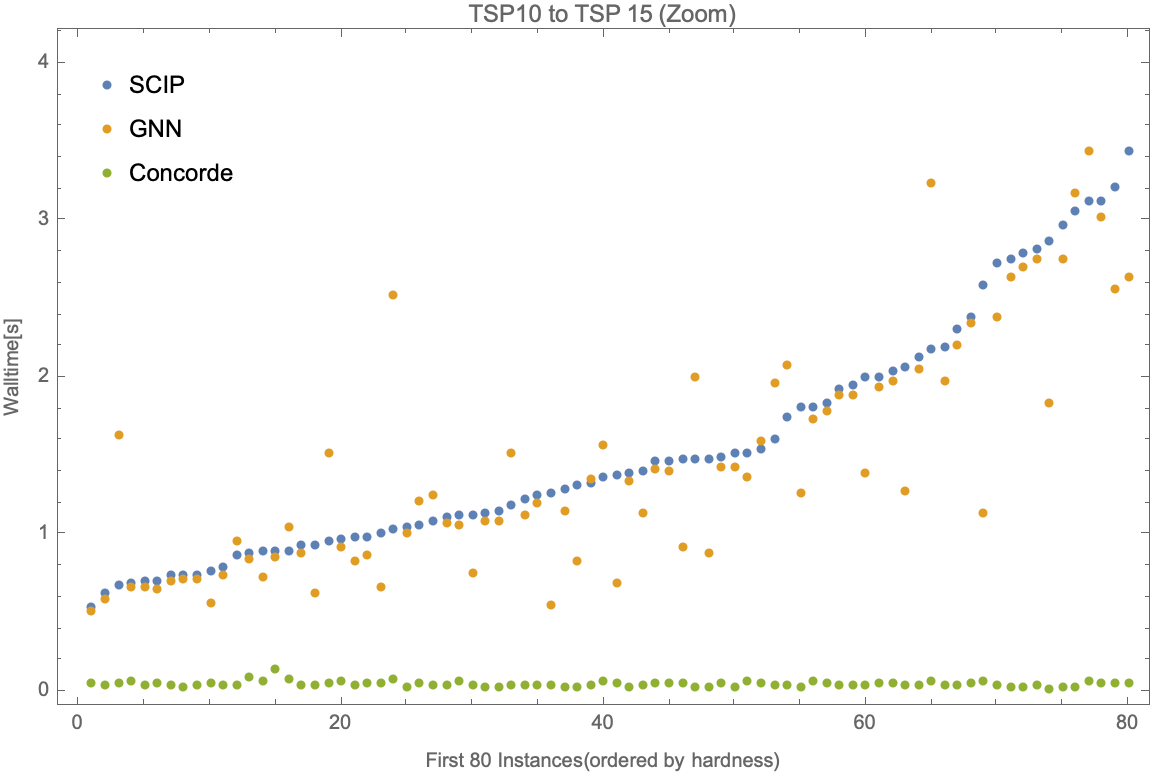}
	\end{minipage}
	\begin{minipage}[b]{0.5\linewidth}
		\centering
		\includegraphics[width=\linewidth]{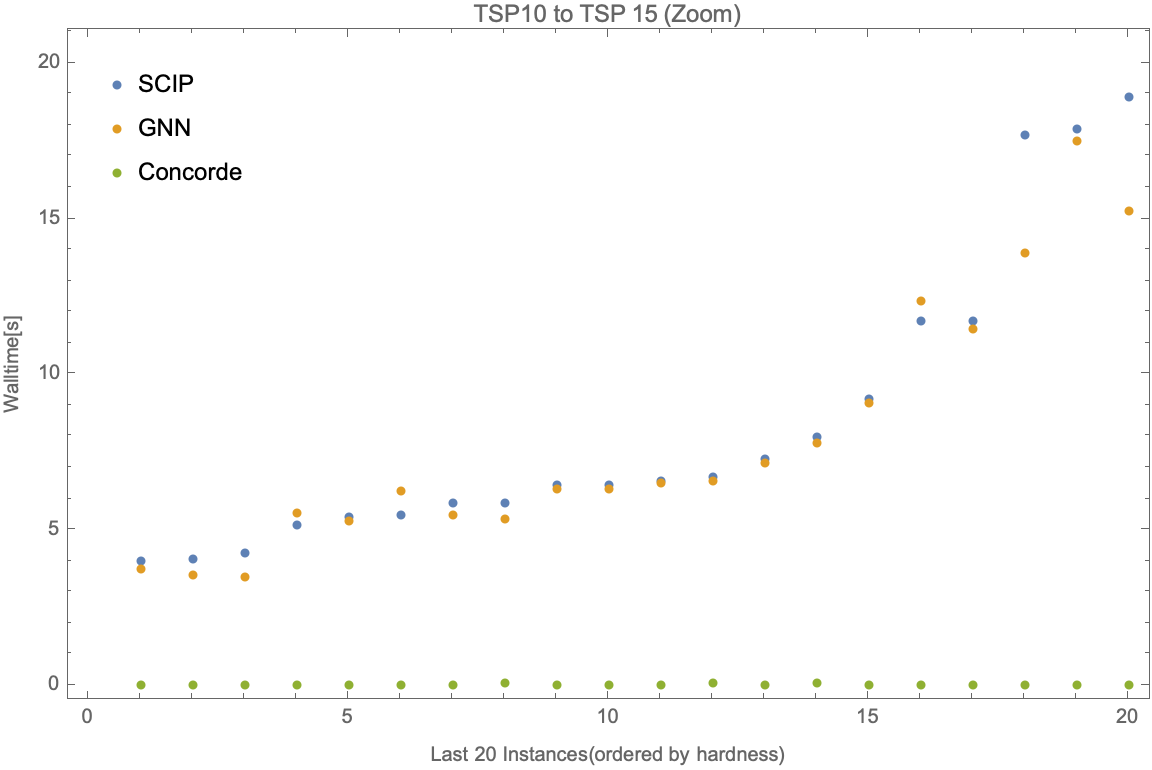}
	\end{minipage}
	\begin{minipage}[b]{0.5\linewidth}
		\centering
		\includegraphics[width=\linewidth]{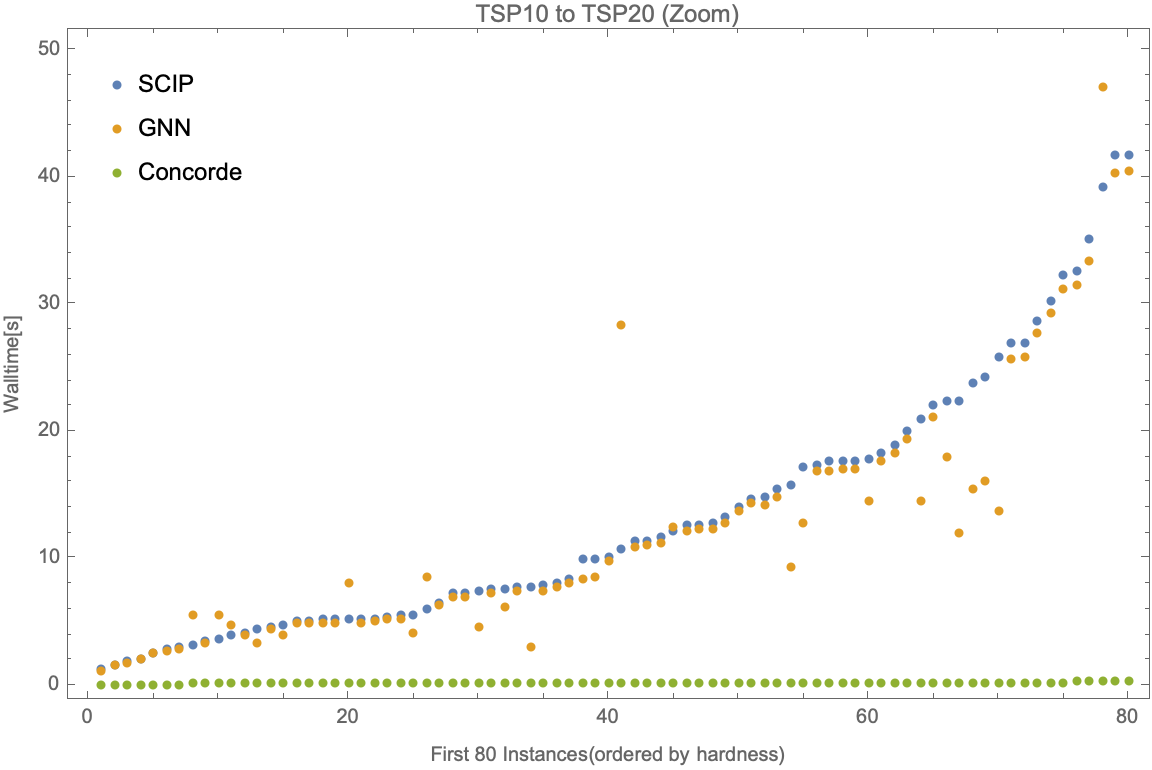}
	\end{minipage}
	\begin{minipage}[b]{0.5\linewidth}
		\centering
		\includegraphics[width=\linewidth]{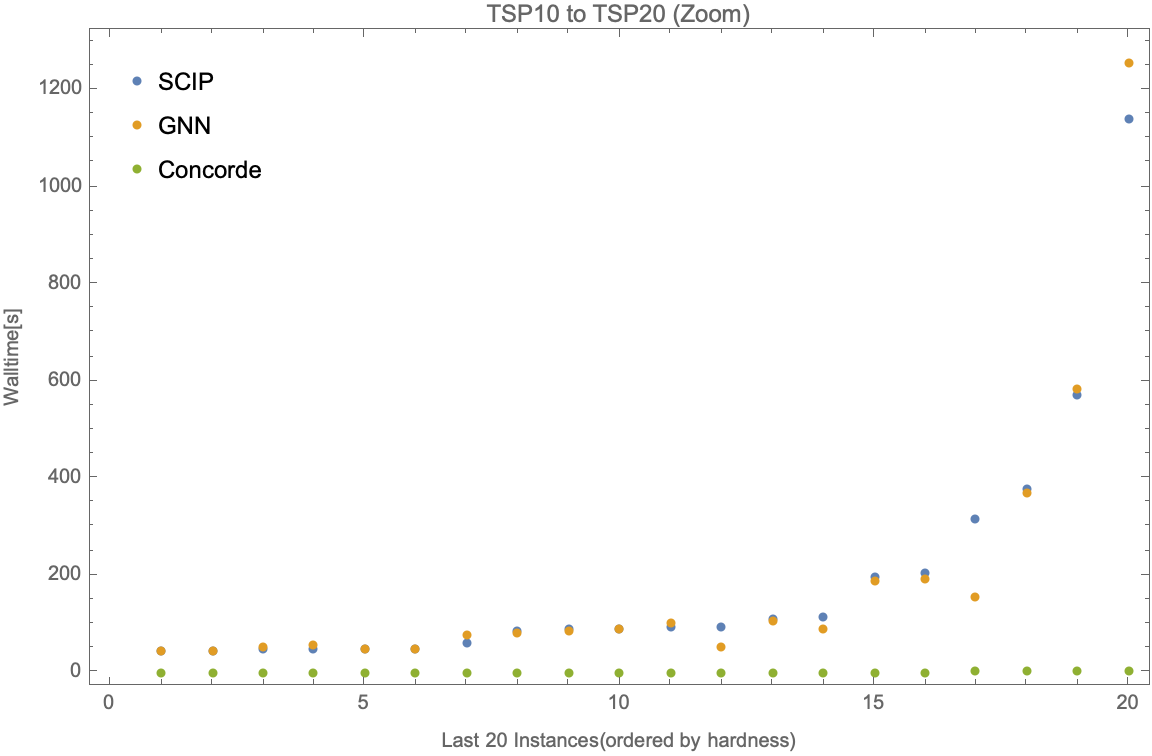}
	\end{minipage}
	\begin{minipage}[b]{0.5\linewidth}
		\centering
		\includegraphics[width=\linewidth]{Figures/results/10-25/zoom-first80.png}
	\end{minipage}
	\begin{minipage}[b]{0.5\linewidth}
		\centering
		\includegraphics[width=\linewidth]{Figures/results/10-25/zoom-last20.png}
	\end{minipage}
	\caption{Result of model trained on TSP10 generalizes to TSP with various sizes with zoomed-in.}
\end{figure}

\subsection{Model Trained on TSP15}\label{apx:tsp15}
\subsubsection{Full Size Plots}
We include the full-size plots for the testing result on the model trained with TSP15.
\begin{figure}[H]
	\centering
	\includegraphics[width=0.8\linewidth]{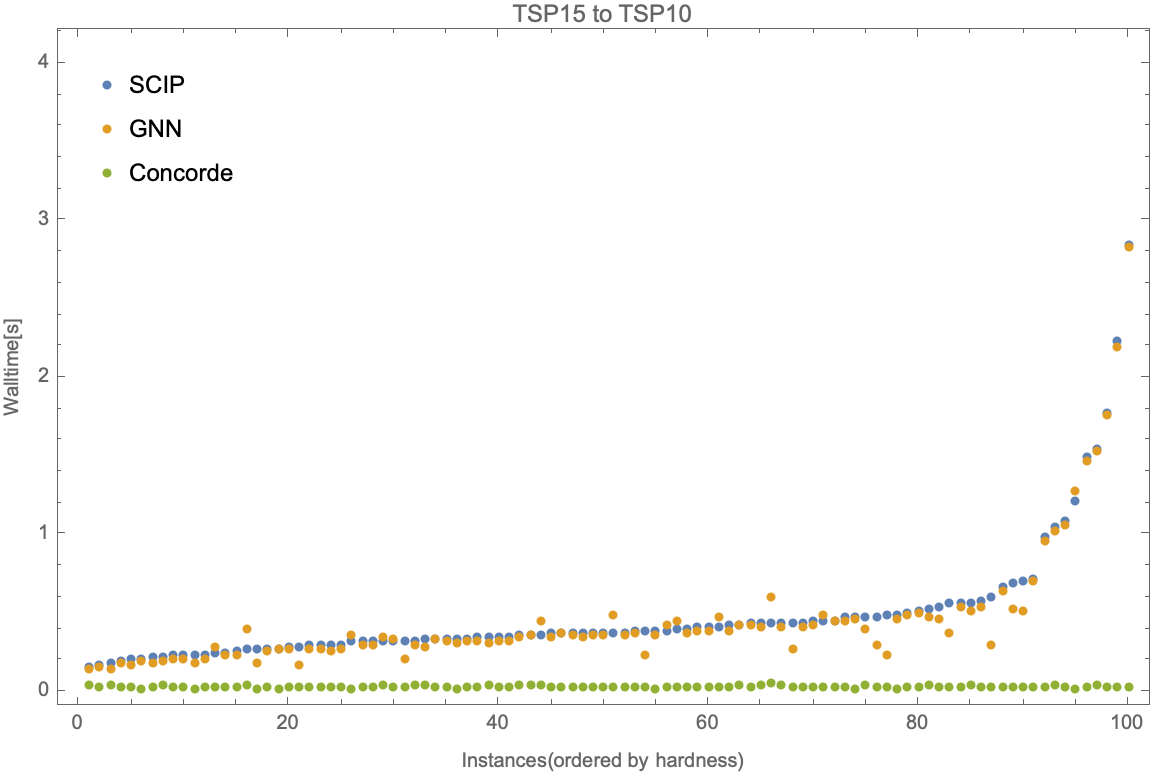}
	\caption{Test on TSP10}
\end{figure}
\begin{figure}[H]
	\centering
	\includegraphics[width=0.8\linewidth]{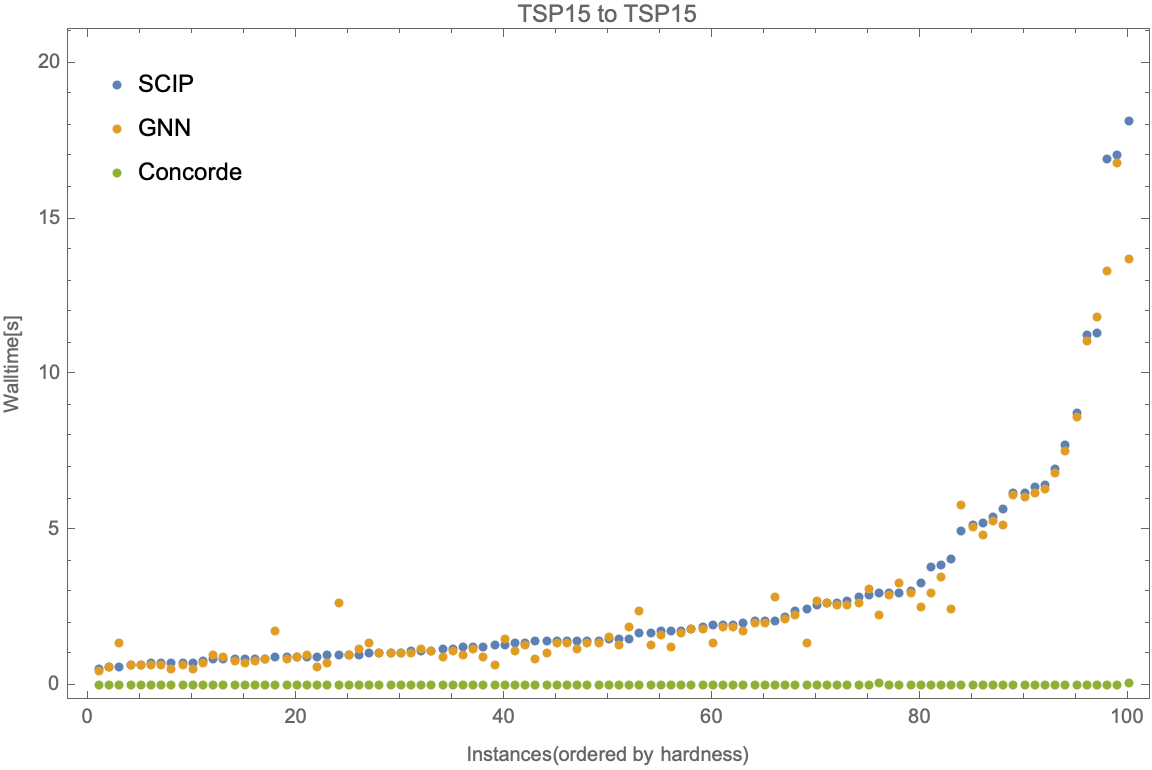}
	\caption{Test on TSP15}
\end{figure}
\begin{figure}[H]
	\centering
	\includegraphics[width=0.8\linewidth]{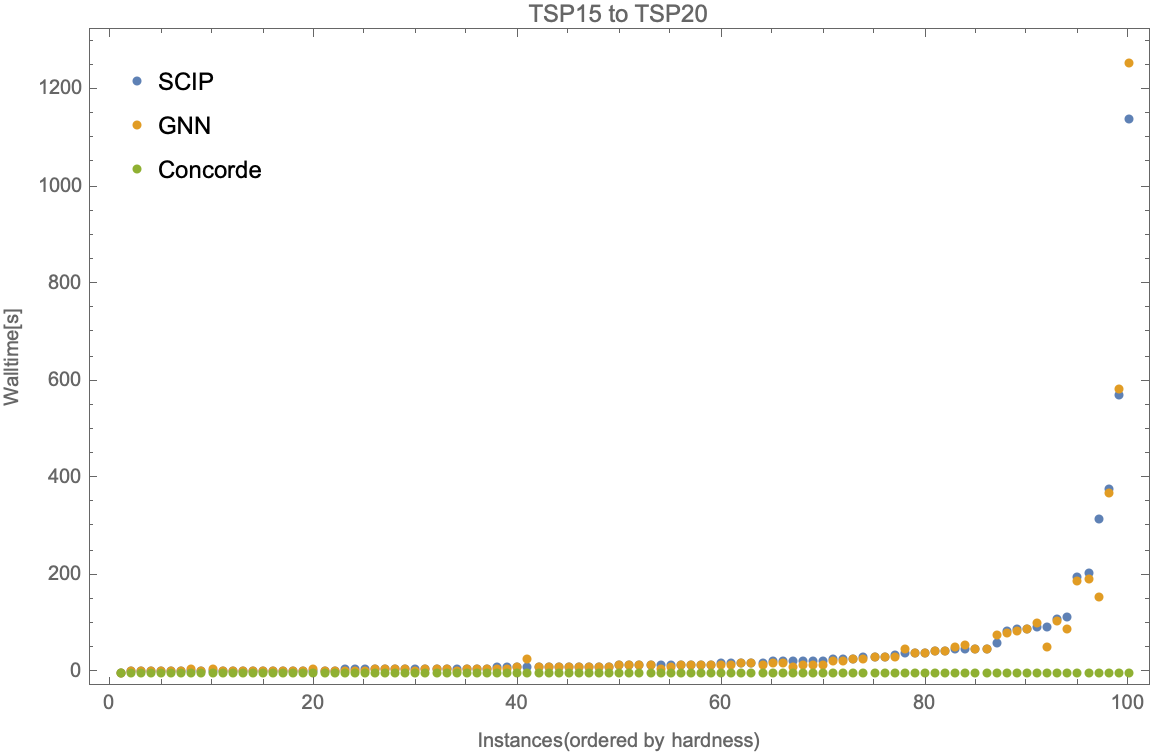}
	\caption{Test on TSP20}
\end{figure}
\begin{figure}[H]
	\centering
	\includegraphics[width=0.8\linewidth]{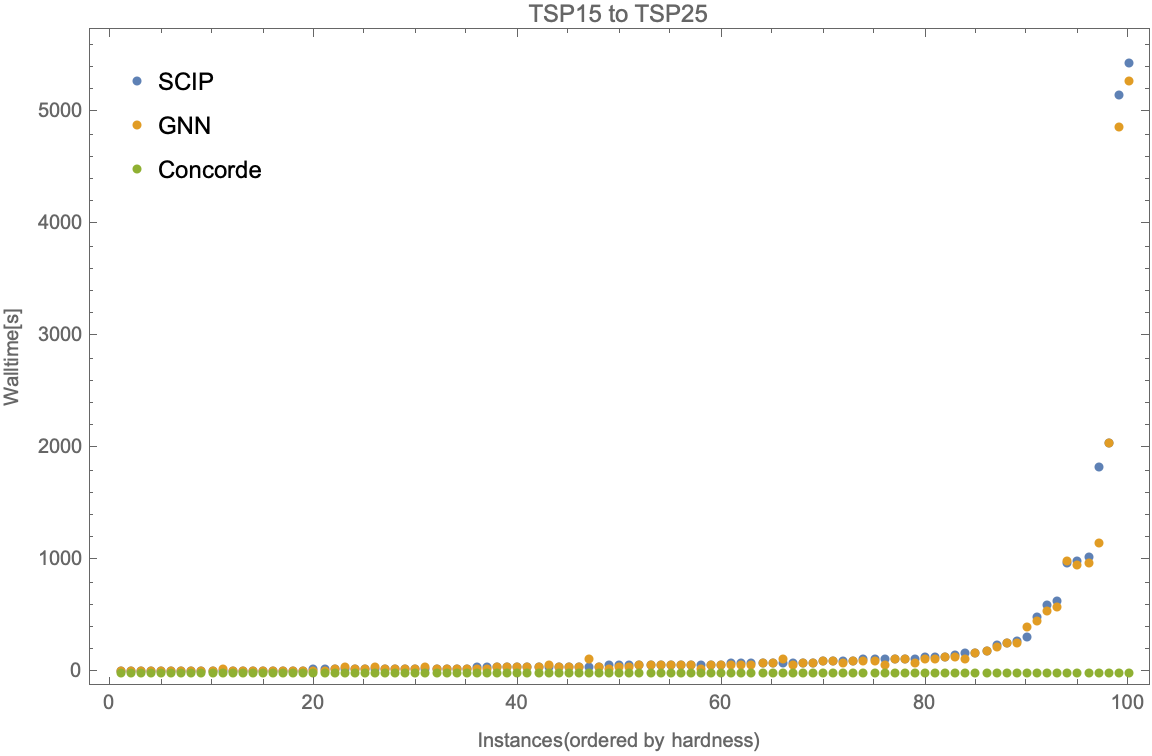}
	\caption{Test on TSP25}
\end{figure}

\subsubsection{Zoom In Plots}
We include the zoom-in plots for the testing result on the model trained with TSP15.
\begin{figure}[H]
	\begin{minipage}[b]{0.5\linewidth}
		\centering
		\includegraphics[width=\linewidth]{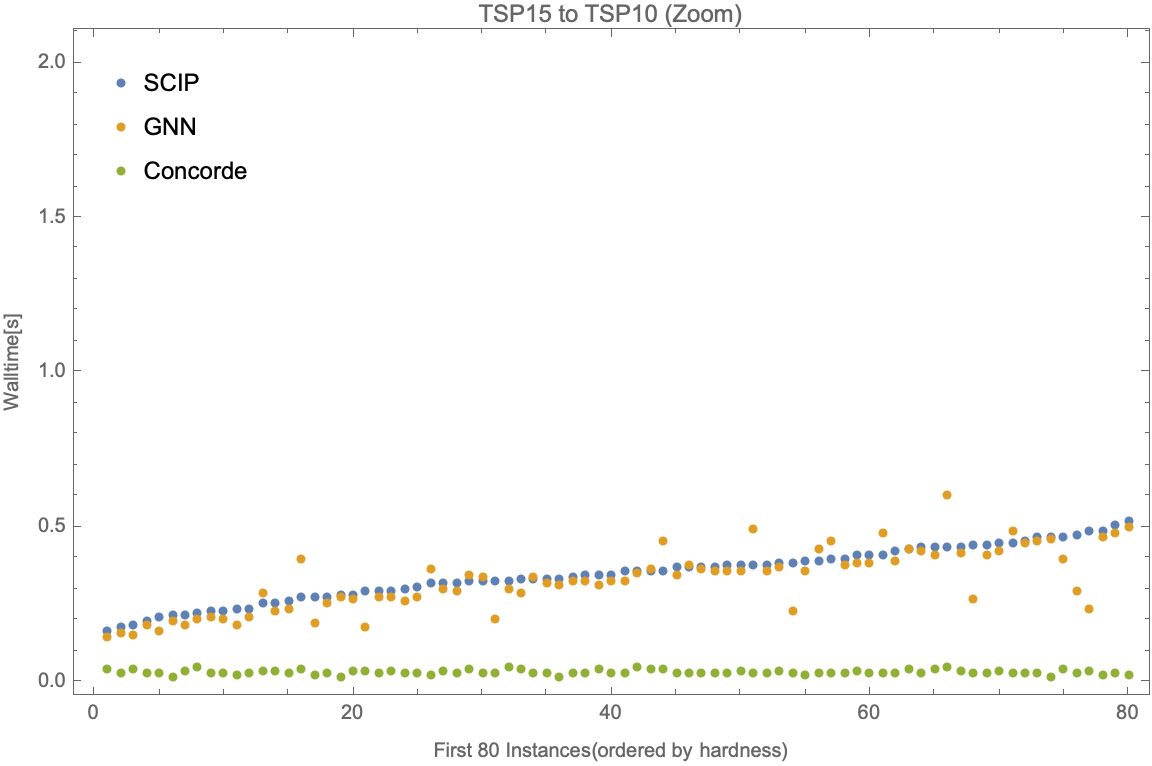}
	\end{minipage}
	\begin{minipage}[b]{0.5\linewidth}
		\centering
		\includegraphics[width=\linewidth]{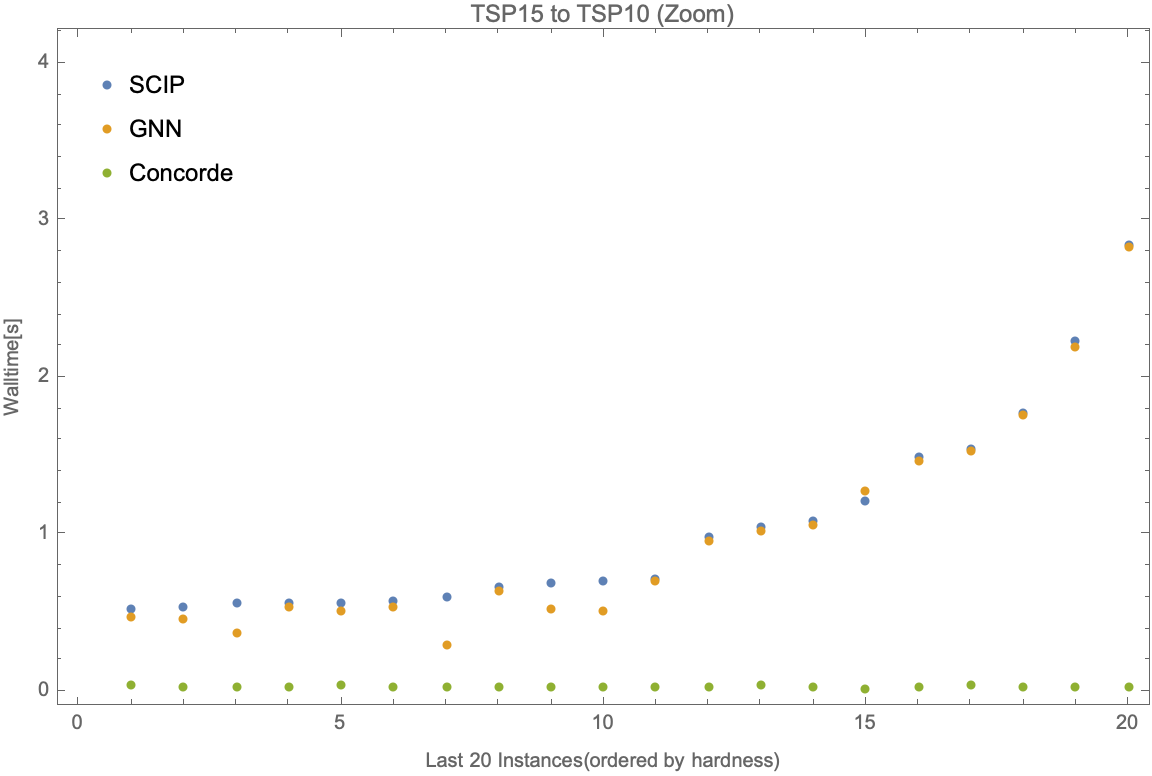}
	\end{minipage}
	\begin{minipage}[b]{0.5\linewidth}
		\centering
		\includegraphics[width=\linewidth]{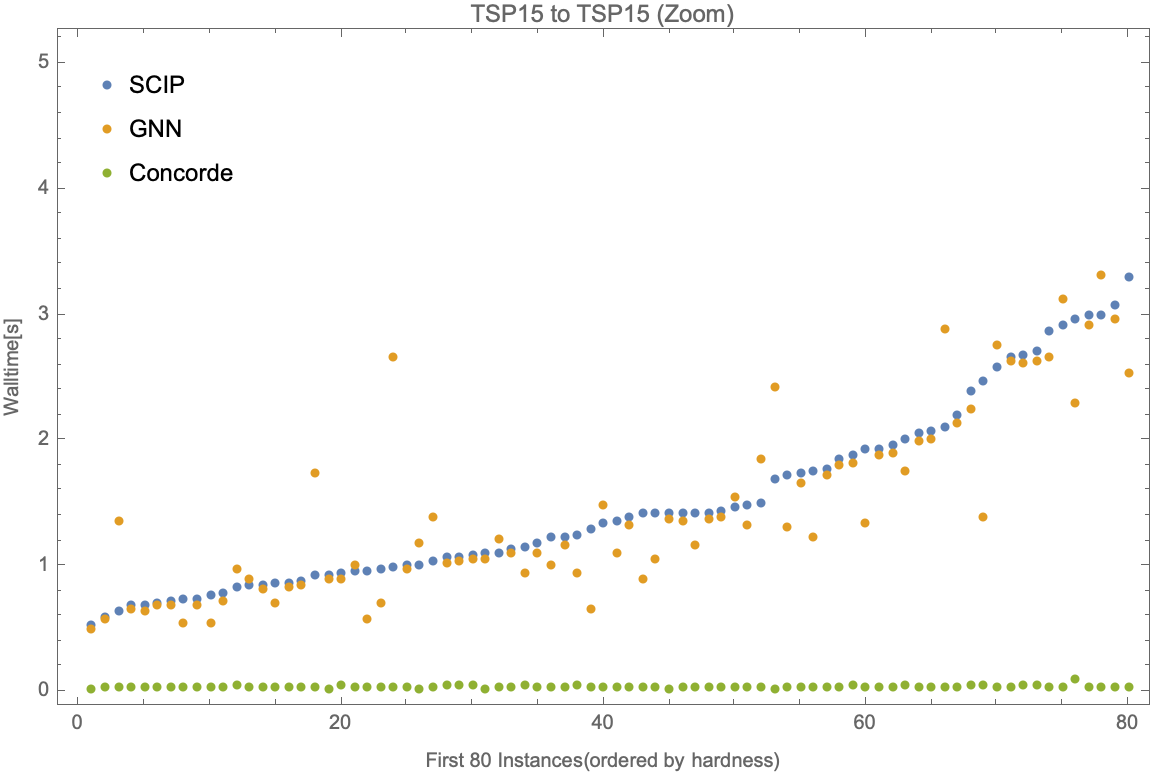}
	\end{minipage}
	\begin{minipage}[b]{0.5\linewidth}
		\centering
		\includegraphics[width=\linewidth]{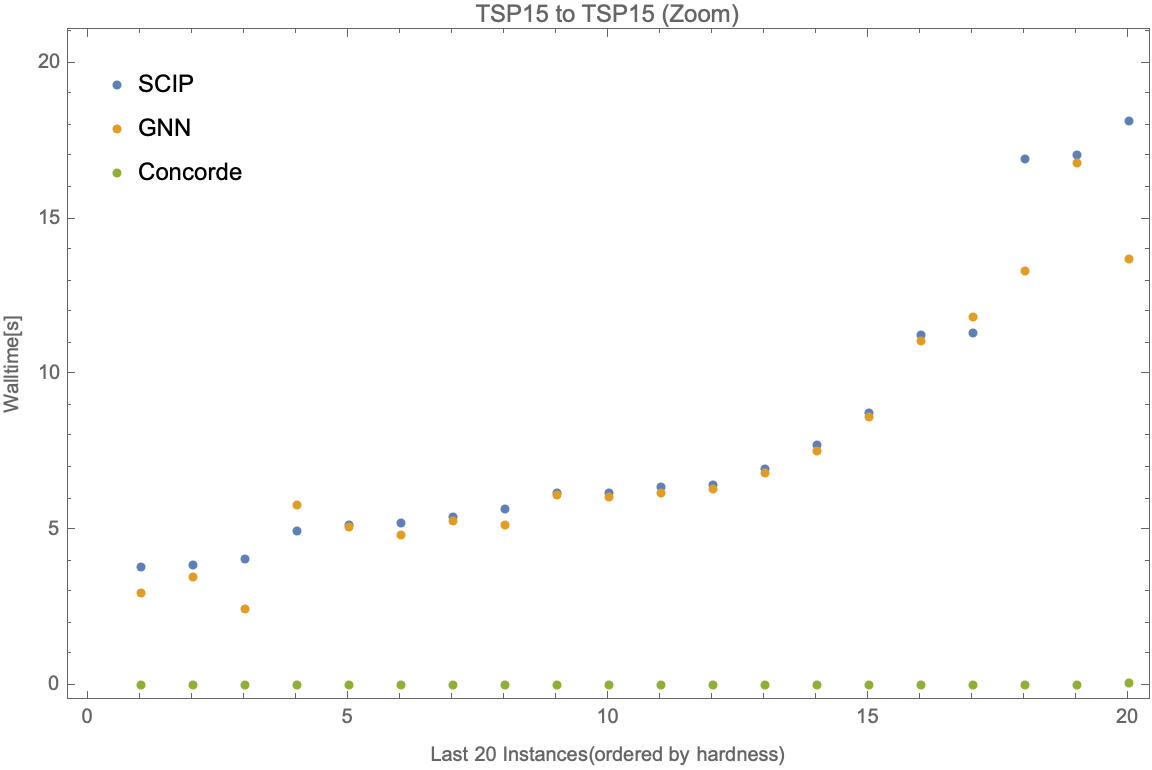}
	\end{minipage}
	\begin{minipage}[b]{0.5\linewidth}
		\centering
		\includegraphics[width=\linewidth]{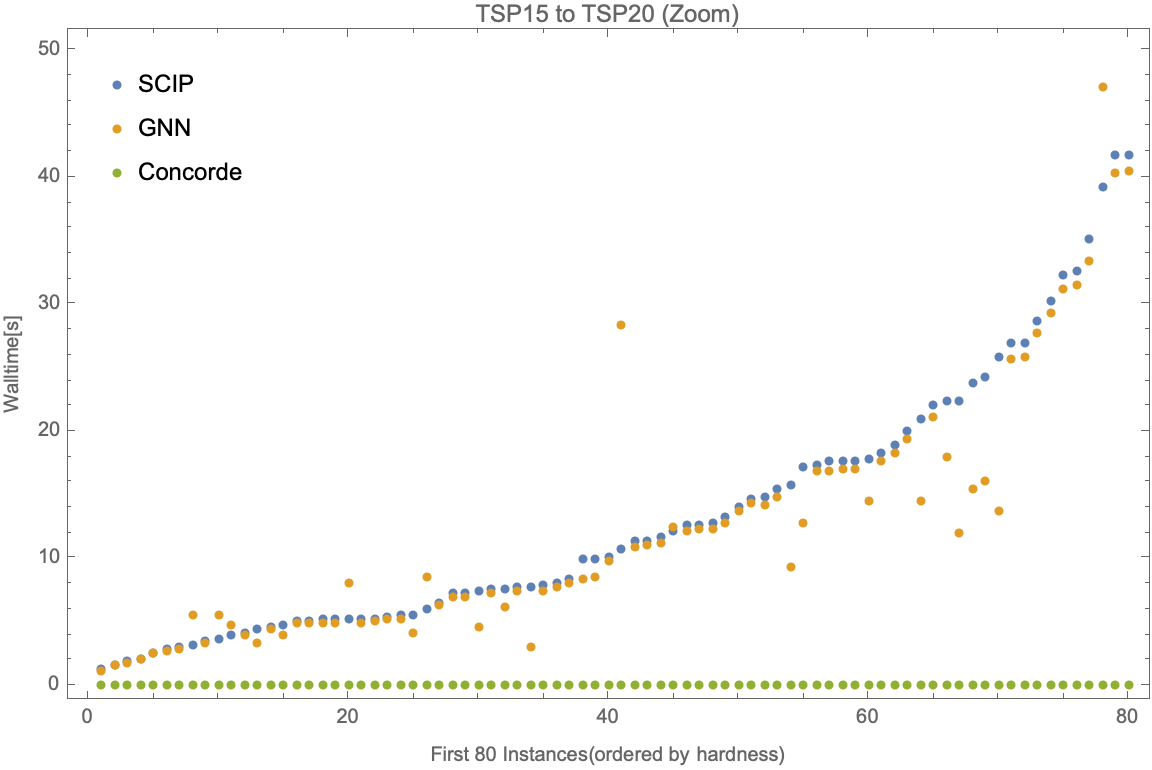}
	\end{minipage}
	\begin{minipage}[b]{0.5\linewidth}
		\centering
		\includegraphics[width=\linewidth]{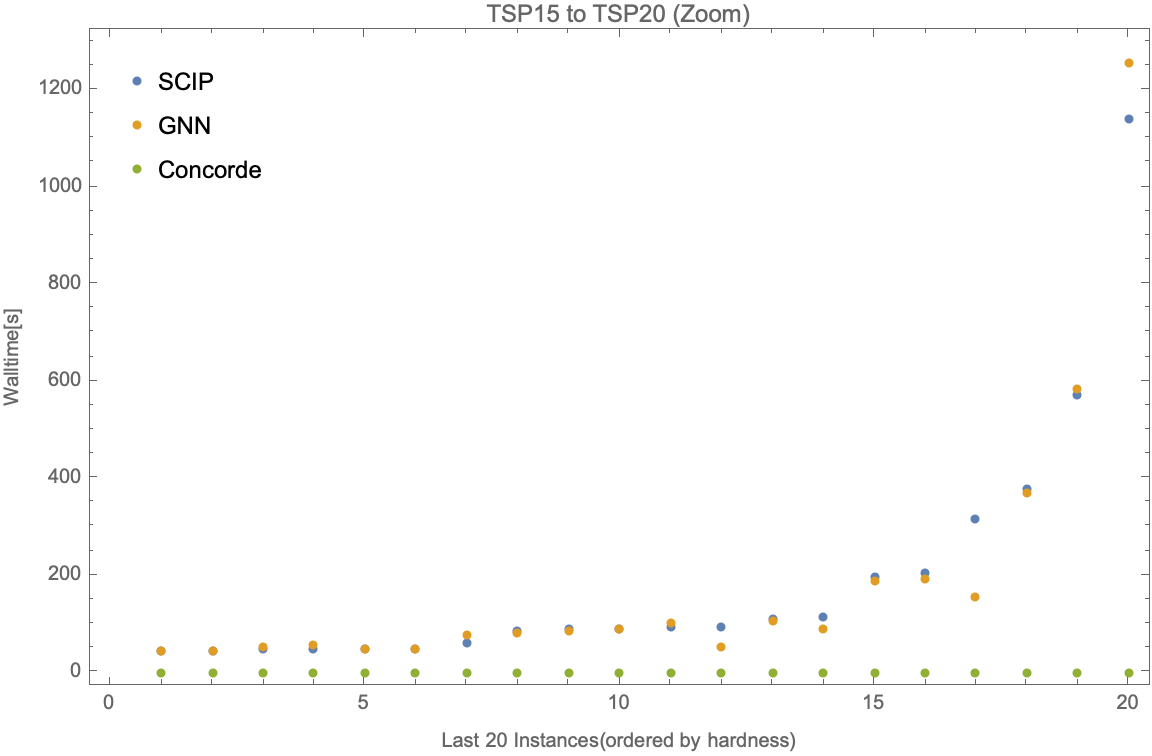}
	\end{minipage}
	\begin{minipage}[b]{0.5\linewidth}
		\centering
		\includegraphics[width=\linewidth]{Figures/results/15-25/zoom-first80.png}
	\end{minipage}
	\begin{minipage}[b]{0.5\linewidth}
		\centering
		\includegraphics[width=\linewidth]{Figures/results/15-25/zoom-last20.png}
	\end{minipage}
	\caption{Result of model trained on TSP15 generalizes to TSP with various sizes with zoomed-in.}
\end{figure}

\subsection{Comparison between \texttt{Gurobi} and \texttt{Concorde}}\label{apx:Concorde_Gurobi}
As a sanity check, we compare the performance between \texttt{Concorde} and the TSP API provided by \texttt{Gurobi}.

\begin{figure}[H]
	\centering
	\includegraphics[width=0.8\linewidth]{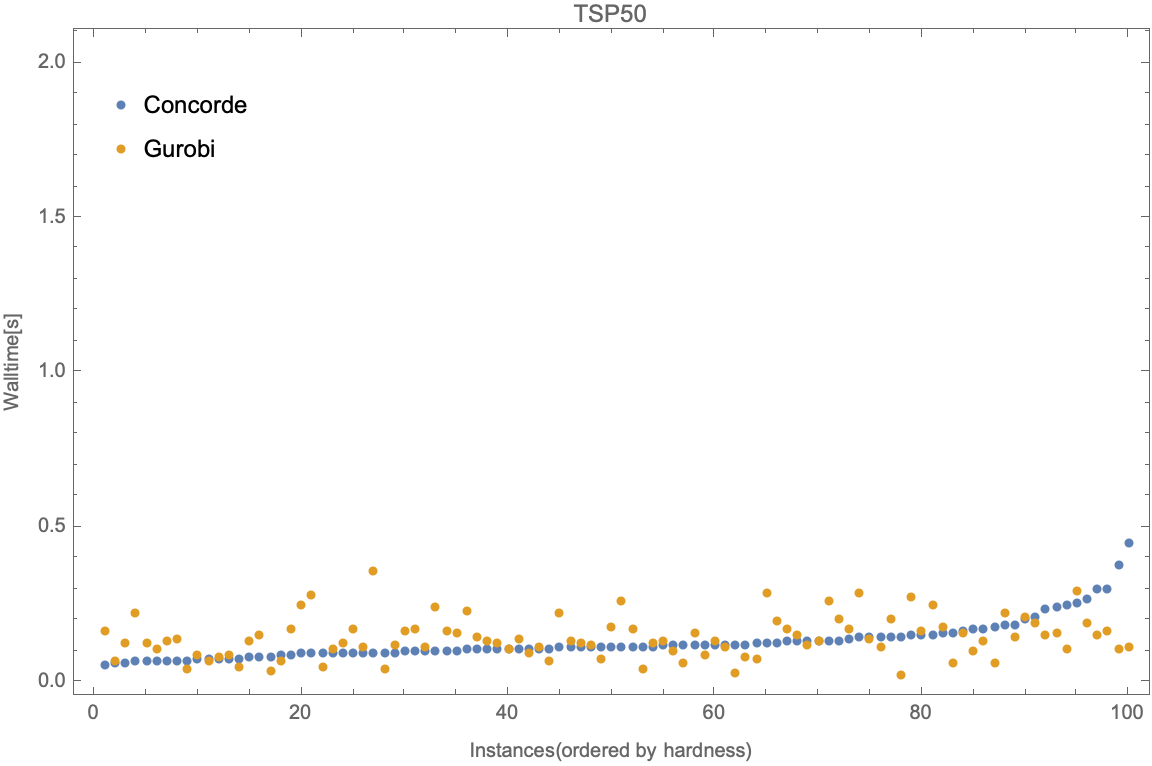}
	\caption{\texttt{Concorde} v.s. \texttt{Gurobi} on TSP50}
	\label{fig:tsp50}
\end{figure}
\begin{figure}[H]
	\centering
	\includegraphics[width=0.8\linewidth]{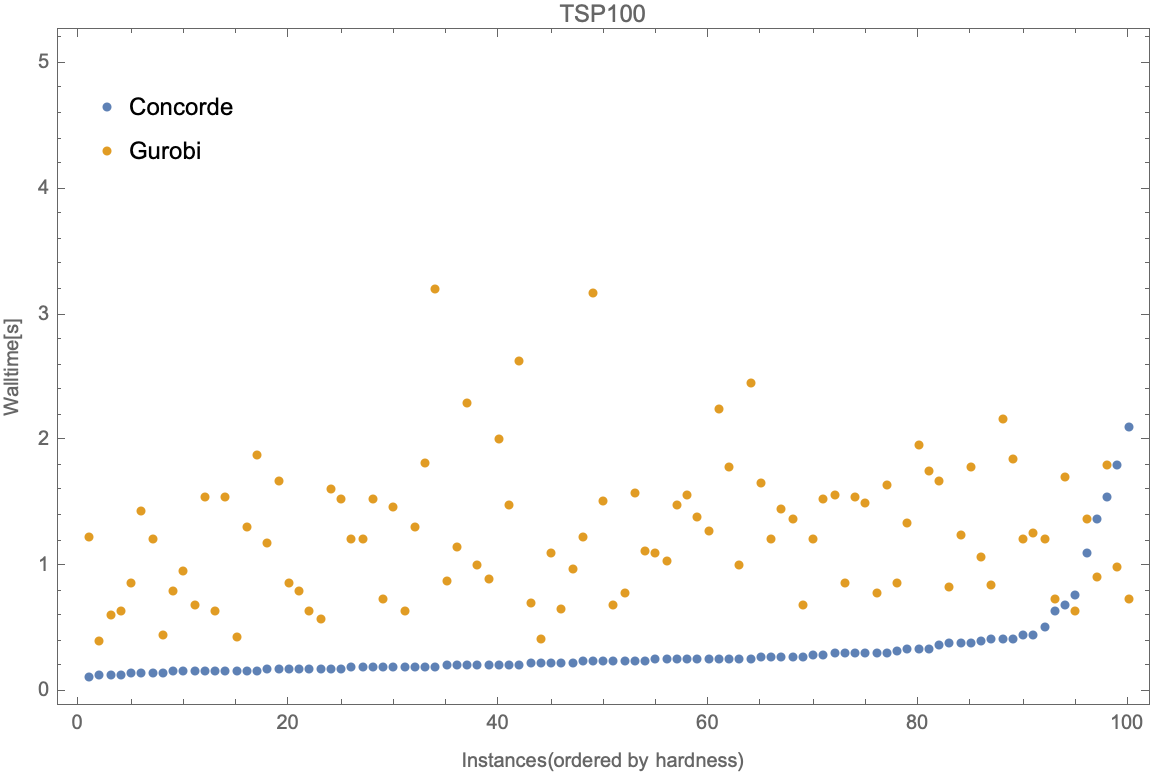}
	\caption{\texttt{Concorde} v.s. \texttt{Gurobi} on TSP100}
	\label{fig:tsp100}
\end{figure}

\end{document}